\documentclass[10pt,twocolumn,letterpaper]{article}

\usepackage[pagenumbers]{cvpr} % To force page numbers, e.g. for an arXiv version

\usepackage{graphicx}
\usepackage{amsmath}
\usepackage{amssymb}
\usepackage{booktabs}

\usepackage[pagebackref,breaklinks,colorlinks]{hyperref}

\usepackage[capitalize]{cleveref}
\crefname{section}{Sec.}{Secs.}
\Crefname{section}{Section}{Sections}
\Crefname{table}{Table}{Tables}
\crefname{table}{Tab.}{Tabs.}
\usepackage[dvipsnames]{xcolor}
\usepackage{multirow}
\usepackage{bbm}

\DeclareMathOperator*{\argmin}{arg\,min}
 
\newcommand*{\ours}{AsyInst}

\def\cvprPaperID{2327} % *** Enter the CVPR Paper ID here
\def\confName{CVPR}
\def\confYear{2023}

\begin{document}

%%%%%%%%% TITLE - PLEASE UPDATE
\title{\ours: Asymmetric Affinity with DepthGrad and Color for Box-Supervised Instance Segmentation}

% \author{Siwei Yang\\
% Tongji University\\
% Institution1 address\\
% {\tt\small firstauthor@i1.org}
% % For a paper whose authors are all at the same institution,
% % omit the following lines up until the closing ``}''.
% % Additional authors and addresses can be added with ``\and'',
% % just like the second author.
% % To save space, use either the email address or home page, not both
% \and
% Second Author\\
% Institution2\\
% First line of institution2 address\\
% {\tt\small secondauthor@i2.org}
% }

\author{
Siwei Yang${^{1}}$
\qquad
Longlong Jing${^{2}}$
\qquad
Junfei Xiao${^{3}}$
\qquad
Hang Zhao${^{4}}$
\qquad
Alan Yuille${^{3}}$
\qquad
Yingwei Li${^{3}}$
\\
\\$^{1}$Tongji University \qquad  \qquad $^{2}$The City University of New York \\$^{3}$Johns Hopkins University \qquad $^{4}$Tsinghua University
}

\makeatletter
\let\@oldmaketitle\@maketitle% Store \@maketitle
\renewcommand{\@maketitle}{
\@oldmaketitle% Update \@maketitle to insert...
\centering

\includegraphics[width=1.0\linewidth]{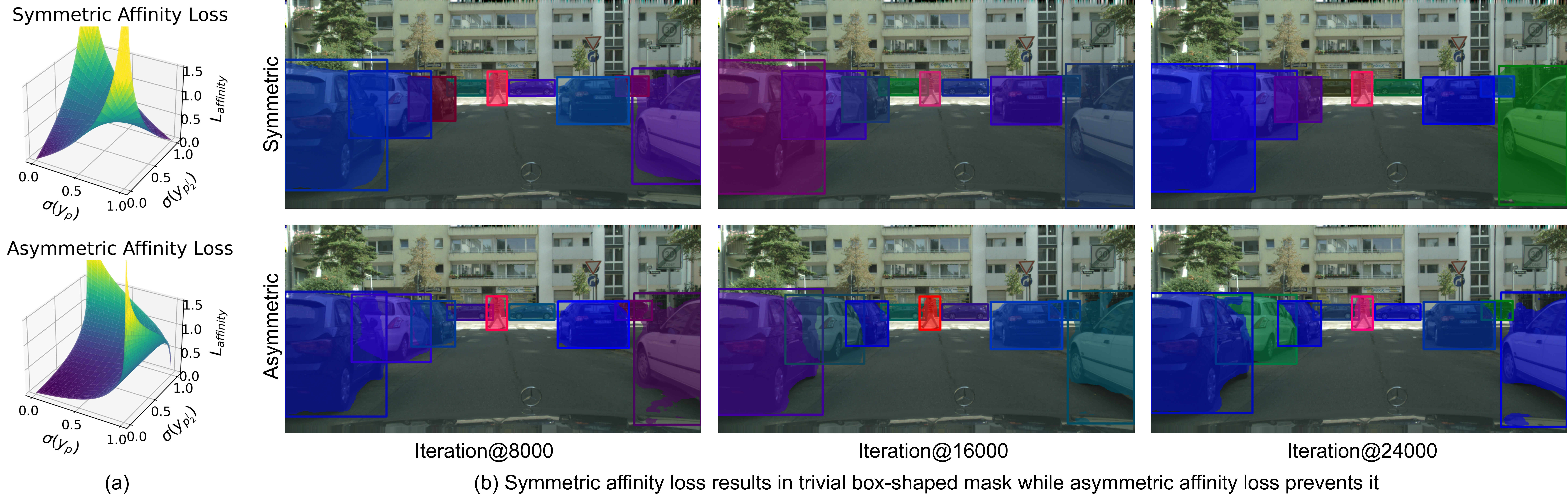}
\captionof{figure}{\textbf{(a)} shows the landscape of symmetric and asymmetric affinity loss in 3D. The definition of notations is explained in \cref{sec:depth}. \textbf{(b)} demonstrates that symmetric affinity loss worsens convergence to trivial box-shaped predictions while the asymmetric affinity loss prevents it. More analysis and qualitative results are discussed in \cref{subsec:exp_asy_trivial}. It should be noted that these results don't represent the final performance of our \ours.}
\label{fig:intro}
\bigskip
}
\makeatother

\maketitle

%%%%%%%%% ABSTRACT
\begin{abstract}
\vspace{-15pt}
The weakly supervised instance segmentation is a challenging task. The existing methods typically use bounding boxes as supervision and optimize the network with a regularization loss term such as pairwise color affinity loss for instance segmentation. Through systematic analysis, we found that the commonly used pairwise affinity loss has two limitations: (1) it works with color affinity but leads to inferior performance with other modalities such as depth gradient, (2)the original affinity loss does not prevent trivial predictions as intended but actually accelerates this process due to the affinity loss term being symmetric. To overcome these two limitations, in this paper, we propose a novel asymmetric affinity loss which provides the penalty against the trivial prediction and generalizes well with affinity loss from different modalities. With the proposed asymmetric affinity loss, our method outperforms the state-of-the-art methods on the Cityscapes dataset and outperforms our baseline method by 3.5\% in mask AP.

\end{abstract}

%%%%%%%%% BODY TEXT
\section{Introduction}
\label{sec:intro}

Instance segmentation is an important yet challenging task in computer vision~\cite{he2017mask,huang2019mask,chen2020blendmask,cheng2020boundary,wang2020solo,tian2020conditional,wang2020solov2,li2017fully,liu2018path,chen2019hybrid,de2017semantic,newell2017associative,liu2017sgn,gao2019ssap,bolya2019yolact}. It requires models to predict the localization of the object and to predict the fine-grained segmentation masks of objects. The training of instance segmentation models requires fine-grained pixel-wised annotations which are expensive and time-consuming to obtain.

To avoid annotating large-scale fine-grained pixel-wised datasets, recently, there are a few attempts to train the model with bounding box labels~\cite{Khoreva_2017_CVPR,NEURIPS2019_e6e71329,tian2021boxinst,li2022box,lan2021discobox}. By training with box-level labels, the cost of training data collection can be greatly reduced. Normally, these methods supervise the model training by using the pairwise loss based on different kinds of inductive prior from images including color similarity~\cite{tian2021boxinst} or bounding box tightness~\cite{NEURIPS2019_e6e71329}. Among all the work, BoxInst~\cite{tian2021boxinst} proposed to use pairwise color affinity loss to optimize the model and achieved promising results. The proposed pairwise color affinity loss enables the model to be learned with pixel-wised fine-level supervision based on the prior knowledge of the color information. Furthermore, this pairwise affinity loss doesn't necessarily depend on color similarity alone and should be capable to be generalized to other modalities as well.

As the existing work mainly relies on RGB images for box-supervised instance segmentation, other modalities are less explored for this task. For example, depth images have a natural advantage in identifying the boundary of objects compared to color information. Therefore, an intuitive idea would be applying the pairwise affinity loss with depth gradient affinity under a simple observation that depth gradient tends to be more consistent in the non-boundary area of objects but only encounters performance drop in mask accuracy. However, we observed that simply adopting the affinity loss on the depth gradient would lead to inferior performance. And our study shows that the original affinity loss being symmetric is the reason behind low performance which not only makes the pairwise affinity loss incompatible with other modalities such as depth gradient but also hurt performance with color affinity.

To understand the limitation of the affinity loss, we visualized the optimization surface of the loss in 3D space in \cref{fig:sym_two_halves}. The saddle-shaped \textit{landscape} of symmetric affinity loss consists of 
two halves \textcolor{Blue}{one} of which converges to produce double-negative pixel pairs whereas the \textcolor{Red}{other one} converges to produce double-positive pixel pairs
In addition to the symmetry, the \textcolor{Red}{double-positive half} of the affinity loss is more effective than the \textcolor{Blue}{other one} under the influence of non-positive gradient from the projection loss~\cite{tian2021boxinst}, which is a widely used method to provide box-level supervision~\cite{tian2021boxinst,li2022box}. This behavior results in trivial predictions such as box-shaped masks. More details related to the interaction between symmetric affinity loss and projection loss can be found in \cref{sec:proj_sym_affinity}.

Therefore, we intend to propose an asymmetric affinity loss to increase the chance of a pixel pair falling on the affinity loss's \textcolor{Blue}{double-negative half} to compensate for the bias towards positive pixels from the projection loss. More specifically, we add an offset $\delta$ into the original symmetric affinity loss which controls the degree of asymmetry of the affinity loss. Although a large $\delta$ increases the asymmetry level, it also causes gradient vanishing, thus leading to slow convergence towards double-negative pixel pairs, which decreases the desired compensating bias. A smoothness hyper-parameter $\gamma$ is then introduced to reshape the loss function and alleviate gradient vanishing. Some qualitative results shown in \cref{fig:intro} demonstrate that the original symmetric affinity loss worsens convergence to trivial box-shaped predictions while our proposed asymmetric affinity loss prevents it.

The proposed asymmetric affinity loss not only works well with depth-grad affinity but also improves the previously proposed symmetric affinity loss with color affinity, demonstrating that the proposed asymmetric affinity loss can be a \textit{general} form of pairwise regularization with both color and depth-grad affinity while the symmetric one can only work with color affinity.

To summarise, the main contribution of this paper is threefold:
\begin{itemize}
    \item We propose a novel approach to improve the box-supervised instance segmentation model with depth information via depth gradient affinity.

    \item Through extensive analysis, we reveal a previously overlooked common flaw in symmetric affinity loss and propose a simple yet effective solution to address it by making it asymmetric. 
    
    \item We conduct thorough and conclusive experiments showing the effectiveness of the proposed asymmetric affinity loss. And the proposed \ours~with both color and depth affinity achieves a total performance improvement of \textbf{3.54\%} compared to our baseline on Cityscapes~\cite{Cordts2016Cityscapes}.

\end{itemize}

% Update the cvpr.cls to do the following automatically.
% For this citation style, keep multiple citations in numerical (not
% chronological) order, so prefer \cite{Alpher03,Alpher02,Authors14} to
% \cite{Alpher02,Alpher03,Authors14}.

%------------------------------------------------------------------------

\section{Related Work}
\label{sec:related_work}

\paragraph{Instance Segmentation:} is a challenging task requiring both instance-level and pixel-level predictions and has attracted increasing attention. The existing work can be classified into three categories. The top-down methods~\cite{li2017fully,he2017mask,liu2018path,chen2019hybrid} tackle instance segmentation by detecting objects first and then predicting masks inside the detected bounding box. The bottom-up methods~\cite{de2017semantic,newell2017associative,liu2017sgn,gao2019ssap} adopt a classify-then-cluster paradigm in which semantic segmentation is first performed on an image before pixels are clustered into objects with pixel-wise embeddings. Some recent methods~\cite{bolya2019yolact,tian2020conditional,chen2020blendmask,wang2020solo,wang2020solov2} follow a hybrid pattern by combining both top-down and bottom-up approaches which achieve high performance with a computational cost comparable to detection models. Although performing pixel-level predictions in addition to instance-level predictions with little computation overhead, instance segmentation models are still not used as widely as object detection models in the industry mainly due to the highly expensive mask annotations, demonstrating the urgency and importance of box-supervised instance segmentation methods.

\paragraph{Box-supervised Instance Segmentation:} The Box-supervised instance segmentation with deep learning has been less explored so far. SDI~\cite{Khoreva_2017_CVPR} is the first instance segmentation model with deep learning and uses GrabCut to generate pseudo segmentation ground-truth providing pixel-wise supervision. BBTP~\cite{NEURIPS2019_e6e71329} formulates the box-supervised instance segmentation as a multiple instance learning problem where the positive and negative bags are sampled according to box annotations on which unary loss is applied. BBTP also utilizes pairwise affinity loss but both the unary loss and pairwise affinity loss are defined only based on box annotation and there is no supervision for segmentation finer than box annotation. BoxInst~\cite{tian2021boxinst} proposes to use projection loss and color affinity loss to supervise segmentation prediction on the box level and pixel-pair level respectively. Despite the proposed color affinity loss appearing to be very effective, its performance is sub-optimal and can be further improved due to its symmetry based on our analysis and experiment results. None of the above mentioned methods use depth information alone with box annotations, which have a natural advantage at identifying objects' boundaries and is cheap to acquire in contrast to actual fine-grained mask annotations.

\section{Depth-Grad Affinity Regularization}
\label{sec:depth}

In this section, we give a formal definition of the depth gradient affinity and a detailed design of a pairwise affinity loss based on this depth gradient affinity.

\subsection{Definition of Depth Gradient Affinity}
\label{subsec:depth_affinity}
\begin{figure}[!t]
    \centering
    \begin{subfigure}{0.33\linewidth}
        \includegraphics[width=\linewidth]{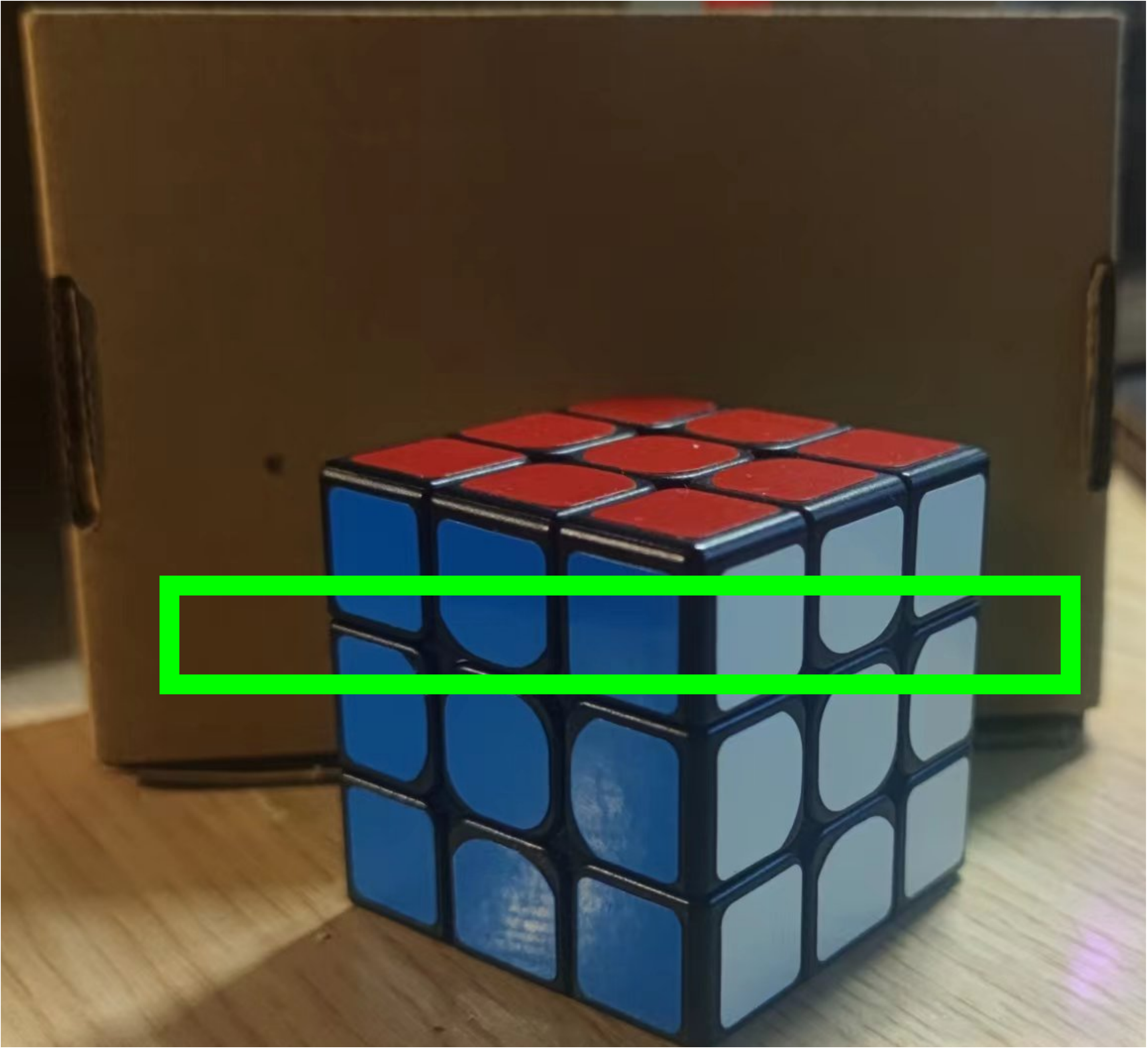}
        % \fbox{\rule{0pt}{2in} \rule{.9\linewidth}{0pt}}
        \caption{An example of possible scenes in real life.}
        \label{fig:depth_grad_demo}
    \end{subfigure}
  \hfill
    \begin{subfigure}{0.66\linewidth}
        % \fbox{\rule{0pt}{2in} \rule{.9\linewidth}{0pt}}
        \includegraphics[width=\linewidth]{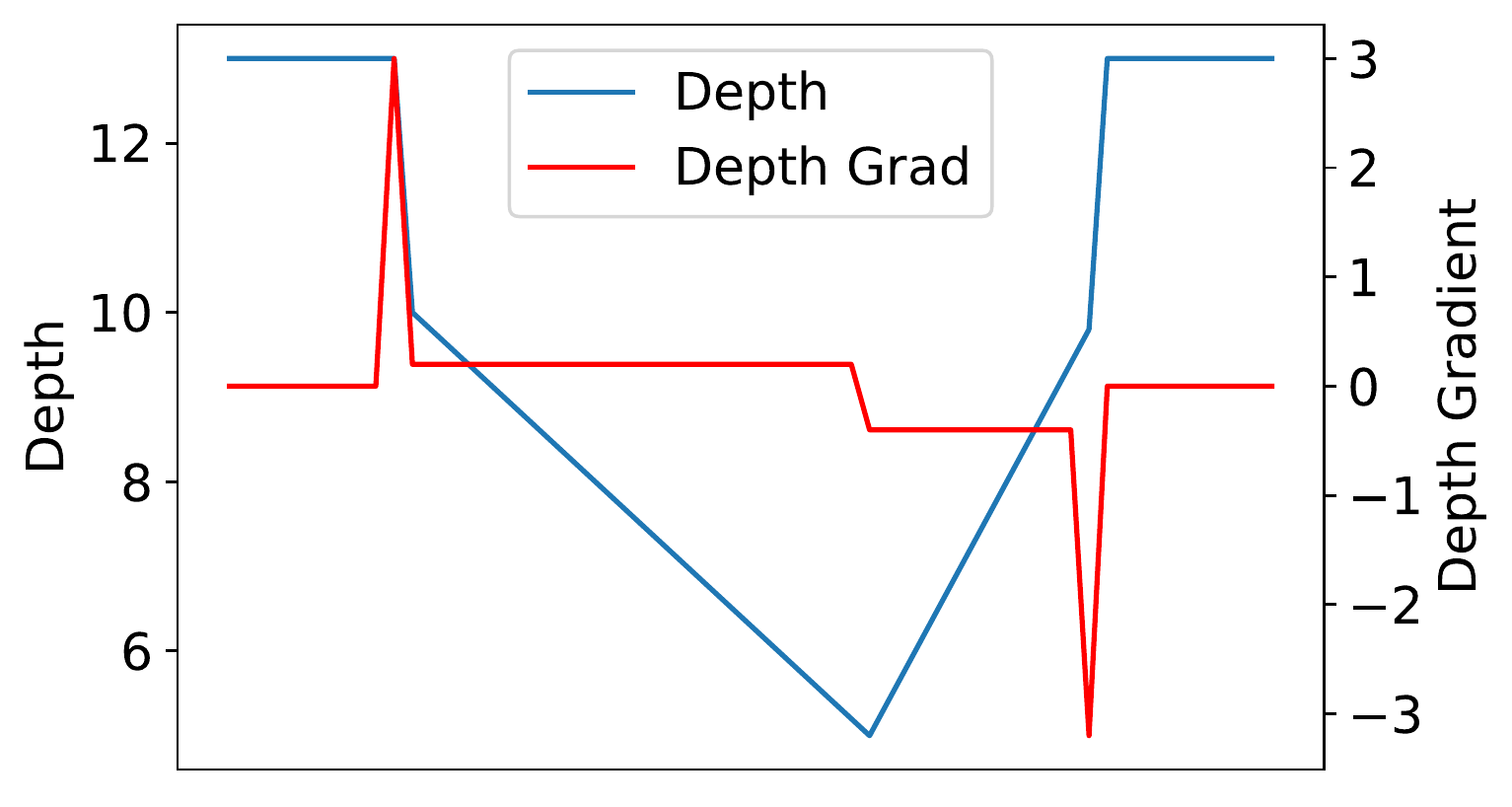}
        \caption{Corresponding depth and depth-grad values.}
        \label{fig:depth_vs_grad}
    \end{subfigure}
    \caption{An example of how depth gradient identifies the boundaries of objects. The depth and horizontal depth-grad values in the green box in \textbf{(a)} is illustrated in \textbf{(b)}.}
    \label{fig:depth_vs_gradient}
    \vspace{-5pt}
\end{figure}

Naturally, depth maps contain more information about objects' shapes and boundaries. As shown in \cref{fig:depth_vs_gradient}, it can be observed that depth gradient (depth-grad) values are more consistent near non-boundary areas. Thus, it is intuitive to assume the pixels with similar depth-grad should have the same pixel-wise mask label and vice versa.

However, such an assumption isn't always applicable in the real world. Pixels on the uneven surface of objects may have different gradient values but still share the same pixel label. For example, as shown in \cref{fig:depth_vs_gradient}, pixels around the cube's edge located inside the green bounding box are less consistent. In other words, the label relationship between  pixels which doesn't share similar depth gradient can be agnostic. Therefore, we only consider the pixels with similar depth gradient values in the definition of depth gradient affinity.

\begin{figure}[!t]
    \centering
    \begin{subfigure}{0.45\linewidth}
        \includegraphics[width=\linewidth]{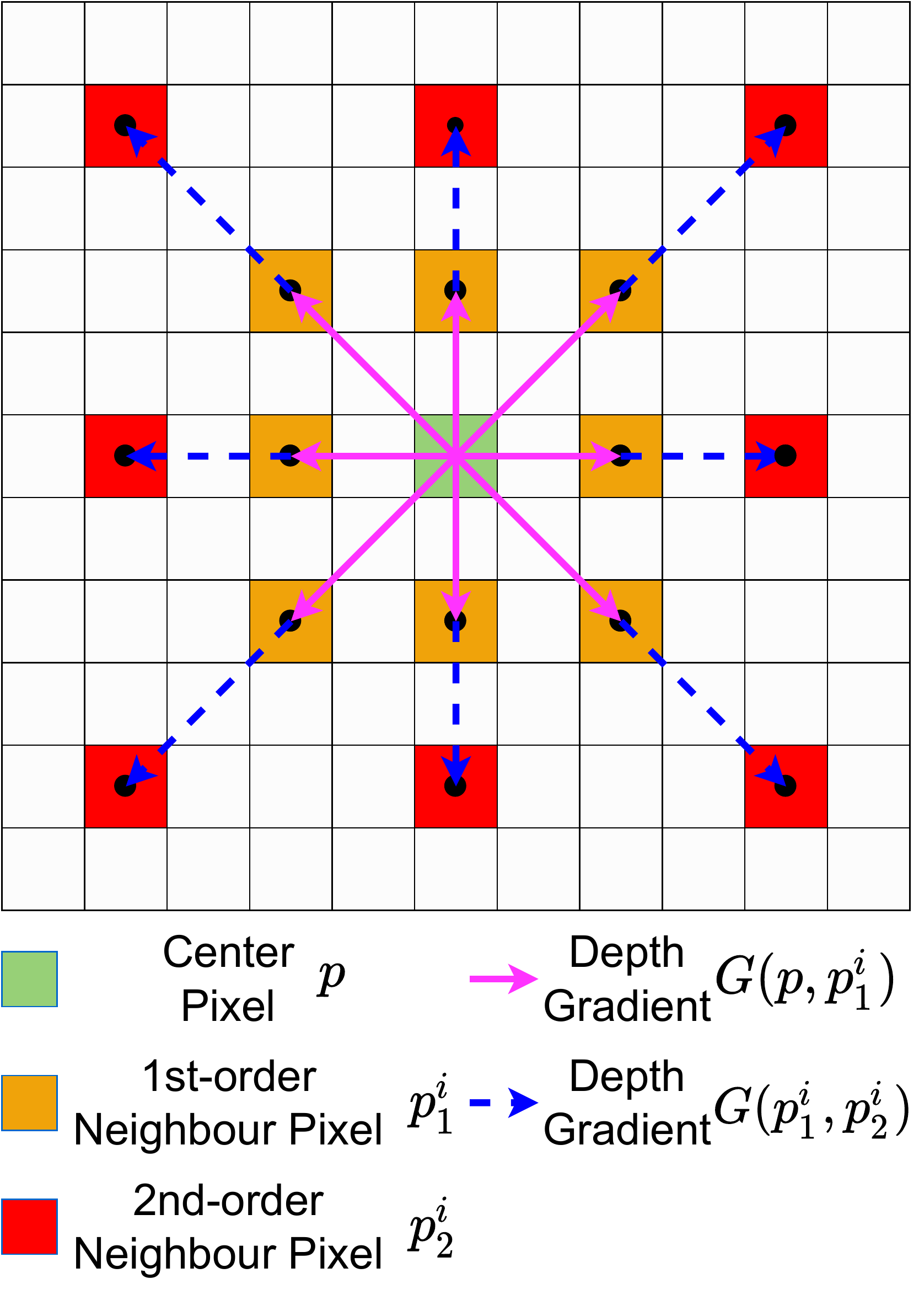}
        \caption{Depth-Grad Affinity}
        \label{fig:depth_grad}
    \end{subfigure}
    \hfill
    \begin{subfigure}{0.54\linewidth}
        \includegraphics[width=\linewidth]{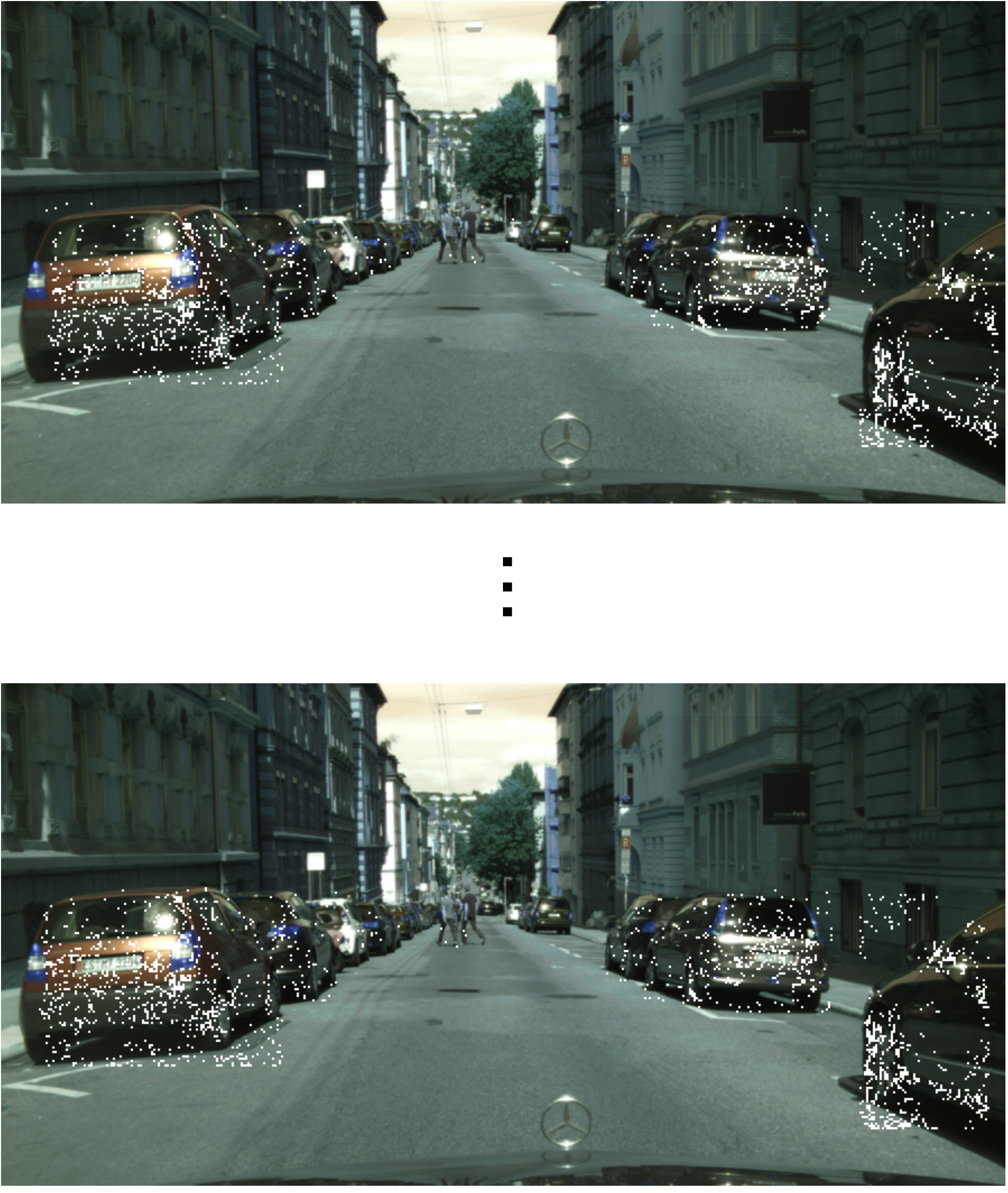}
        \caption{Example depth-grad affinity maps.}
        \label{fig:affinity_map_example}
    \end{subfigure}
    \caption{\textbf{(a)} illustrates the definition of depth-grad affinity (dilation=2) where we compare the depth gradient between the center pixel $p$ and its first order neighbour pixel $p_1^i$ with the depth gradient between the 1st order neighbour pixel $p_1^i$ and the corresponding second order neighbour pixel $p_2^i$. If two gradients are close enough, then $p$ and $p_2^i$ should have the same mask label. \textbf{(b)} shows two examples of all eight depth-grad affinity maps.}
    \label{fig:depth_grad_exmaple}
    \vspace{-5pt}
\end{figure}

Considering an arbitrary pixel $p$ in the image connected with $K \times (K - 1)$ first-order neighbours $p_1^i$ while each has a corresponding second-order neighbour $p_2^i$ as shown in \cref{fig:depth_grad}, the depth gradient between $p$ and $p_1^i$ is denoted as $G(p, p_1^i) = d_{p} - d_{p_1^i}$ while $G(p_1^i, p_2^i) = d_{p_1^i} - d_{p_2^i}$ where $d_{p}, d_{p_1^i}, d_{p_2^i}$ are respectively the depth values of $p, p_1^i, p_2^i$. Then an edge $e$ connecting $p$ and $p_2^i$ is determined as a positive edge in which $p$ and $p_2^i$ have the same label when the difference between $G(p, p_1^i)$ and $G(p_1^i, p_2^i)$ is smaller than a threshold $\tau_\text{d}$, \ie
   
\begin{equation}
    \begin{split}
        \text{Diff}(e) &= |G(p, p_1^i) - G(p_1^i, p_2^i)| \\
        &= |(d_{p} - d_{p_1^i}) - (d_{p_1^i} - d_{p_2^i})| \\
        &= |d_{p} + d_{p_2^i} - 2 * d_{p_1^i}| \\
        &\leq \tau_\text{d}
    \end{split}
    \label{equ:depth_grad}
\end{equation} 
where $\text{Diff}(e)$ is the gradient difference of the edge $e$ and $\tau_\text{d}$ is the threshold for gradient difference

\subsection{Depth-Grad Affinity Loss}

Similar to the color affinity loss of BoxInst~\cite{tian2021boxinst}, we only consider the situation that $\text{Diff}(e) \leq \tau_\text{d}$ in which the edge's label should be $Y_e = 1$ since the $Y_e$ can be agnostic when $\text{Diff}(e) > \tau_\text{d}$ due to the reasons discussed in \cref{subsec:depth_affinity}.

As $e$ being a positive edge implies that $p$ and $p_2^i$ linked by $e$ have the same label, the probability of $Y_e = 1$ is

\begin{equation}
    P(Y_e = 1) = \sigma(y_p) \sigma(y_{p_2^i}) + \sigma(-y_p)\sigma(-y_{p_2^i})
    \label{equ:edge_prob}
\end{equation}

where the $y_p$ and $y_{p_1^i}$ are the logits of pixels $p$ and $p_2^i$, and $\sigma$ the sigmoid function meaning that $\sigma(y_p)$ and $\sigma(y_{p_2^i})$ are the probability of $p$ and $p_2^i$ being the foreground pixel while $\sigma(-y_p)$ and $\sigma(-y_{p_2^i})$ are the probability of $p$ and $p_2^i$ being the negative pixel.

Thus, the depth-grad affinity loss is defined as

\begin{equation}
% \begin{split}
    L_\text{depth} = -\frac{1}{N} \sum_{e \in E_\text{d}^\text{in}}^{\text{Diff}(e) \leq \tau_\text{d}}{\log{P(Y_e = 1)}}
% \end{split}
\label{equ:sym_depth}
\end{equation}

where $E_\text{d}^\text{in}$ is the set of edges where the corresponding $p, p_1^i, p_2^i$ all have a valid depth value and at least one end in the bounding box, and $N$ is the number of edges in $E_\text{d}^\text{in}$ which also satisfy $\text{Diff}(e) \leq \tau_\text{d}$. The purpose of only using edges in $E_\text{d}^\text{in}$ is to focus the regularization inside the box since projection loss can already provide sufficient regularization outside the box.

\section{Coupling between Projection Loss and Symmetric Affinity loss}
\label{sec:proj_sym_affinity}

The proposed depth gradient affinity loss is very reasonably defined based on the experience of color affinity loss proposed in BoxInst~\cite{tian2021boxinst}. However, naively adopting this fashion of pairwise affinity loss with depth gradient affinity has a negative impact on model performance. As illustrated in \cref{fig:intro}, the models trained with depth gradient affinity loss tend to produce trivial box-shaped masks. This is due to that the symmetric property of pairwise affinity loss leads to harmful coupling interaction between the projection loss~\cite{tian2021boxinst} and the pairwise affinity loss.

\subsection{Preference of $L_\text{proj}$ towards False Positive Pixels}
\label{subsec:proj}

The projection loss, originally proposed by Boxinst~\cite{tian2021boxinst}, requires the projection of a predicted probability score mask $m \in {(0, 1)}^{H \times W}$ to be aligned with the corresponding bounding box horizontally and vertically. The horizontal and vertical projection of $m$ are acquired with a $\text{max}$ operation along each axis, \ie

\begin{equation}
    l_x = \text{max}_{y}(m), l_y = \text{max}_{x}(m)
    \label{equ:max_proj}
\end{equation}
where $l_x \in {(0, 1)}^W, l_y \in {(0, 1)}^H$ are the horizontal and vertical projection of $m$ respectively.

Similarly, we can obtain the horizontal and vertical projection of a bounding box noted as $\tilde{l}_x \in {(0, 1)}^W, \tilde{l}_y \in {(0, 1)}^H$, which will be used as ground-truth for $l_x, l_y$. Thus, the projection loss~\cite{tian2021boxinst} is defined as

\begin{equation}
    L_\text{proj} = L_\text{Dice}(l_x, \tilde{l}_x) + L_\text{Dice}(l_y, \tilde{l}_y)\\
    \label{equ:proj_loss_2term}
\end{equation}
where $L_\text{Dice}$ is the Dice loss defined as
\begin{equation}
    \begin{split}
        L_\text{Dice}(l_x, \tilde{l}_x) = 1 - \frac{l_x \cdot \tilde{l}_x}{|l_x|^2 + |\tilde{l}_x|^2}\\
        L_\text{Dice}(l_y, \tilde{l}_y) = 1 - \frac{l_y \cdot \tilde{l}_y}{|l_y|^2 + |\tilde{l}_y|^2}\\
    \end{split}
\end{equation}

Considering the horizontal term of the projection loss $\nabla_{l_x} L_\text{Dice}(l_x, \tilde{l}_x)$ in~\cref{equ:proj_loss_2term}, the gradient of $L_\text{Dice}(l_x, \tilde{l}_x)$ to $l_x$ is

\begin{equation}
    \begin{split}
        \nabla_{l_x} L_\text{Dice}(l_x, \tilde{l}_x) = \frac{2 (l_x \cdot \tilde{l}_x) l_x - (|l_x|^2 + |\tilde{l}_x|^2)\tilde{l}_x}{(|l_x|^2 + |\tilde{l}_x|^2)^2}\\
    \end{split}
\end{equation}

For a pixel in the horizontal projection with the probability score $l_x^k$ and the corresponding ground-truth score $\tilde{l}_x^k$, the $\tilde{l}_x = 1$ if the pixel is horizontal inside the bounding box and $\tilde{l}_x = 0$ if otherwise.

The gradient of $L_\text{Dice}(l_x, \tilde{l}_x)$ to $l_x^k$ is
\begin{equation}
    \frac{\partial L_\text{Dice}(l_x, \tilde{l}_x)}{\partial l_x^k} =\left\{
    \begin{split}
        \frac{2 (l_x \cdot \tilde{l}_x) l_x}{(|l_x|^2 + |\tilde{l}_x|^2)^2} \geq 0 & & \tilde{l}_x = 0\\
        \frac{-(l_x - \tilde{l}_x)^2}{(|l_x|^2 + |\tilde{l}_x|^2)^2} \leq 0 & & \tilde{l}_x = 1\\
    \end{split}
    \right.
\end{equation}

The horizontal projection $l_x$ of the predicted mask $m$ is acquired through $\text{max}$ operation based on~\cref{equ:max_proj} meaning that all the elements in the gradient $\nabla_{m} l_x$ are non-negative. We can infer that the gradient of $L_\text{Dice}(l_x, \tilde{l}_x)$ to any probability scores from the in-box part of $m$ is non-positive while vice versa for probability scores from the out-box part of $m$.

\begin{figure}[!t]
    \centering
    \includegraphics[width=\linewidth]{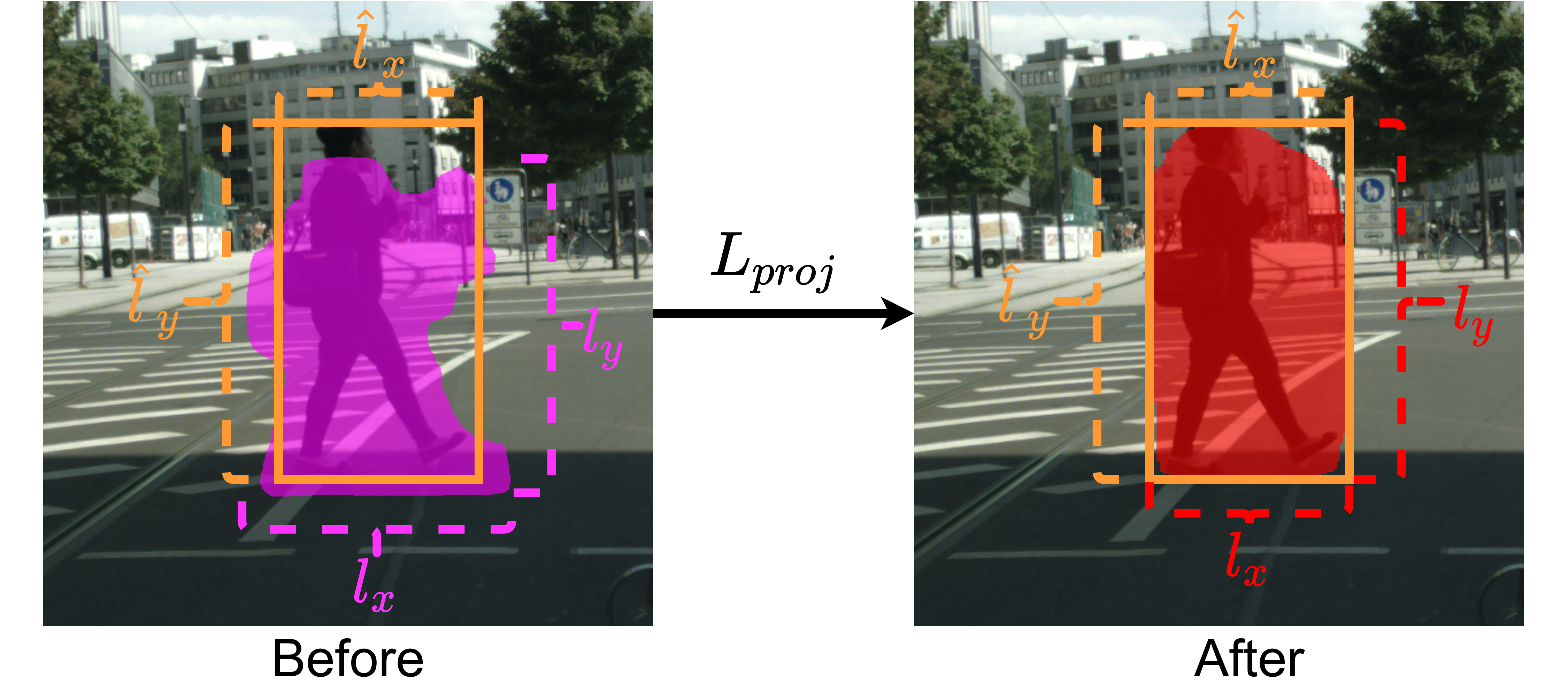}
    \caption{The projection loss aligns the mask and bounding box horizontally and vertically, which leads to bbox-shaped masks.}
    \label{fig:projection_loss}
\end{figure}

As a similar conclusion applies for the gradient of the vertical term in~\cref{equ:proj_loss_2term}, the pixels inside the box can only receive a non-positive gradient from the projection loss which tends to increase the probability scores while vice versa for the pixels outside the box. Thus, the projection loss encourages false positives inside the box leading to trivial predictions such as box-shaped masks as shown in \cref{fig:projection_loss}.

\subsection{Problem with Symmetric Affinity Loss}
\label{subsec:sym_affinity}

\begin{figure}[!h]
    \centering
    \begin{subfigure}{0.49\linewidth}
        \includegraphics[width=\linewidth]{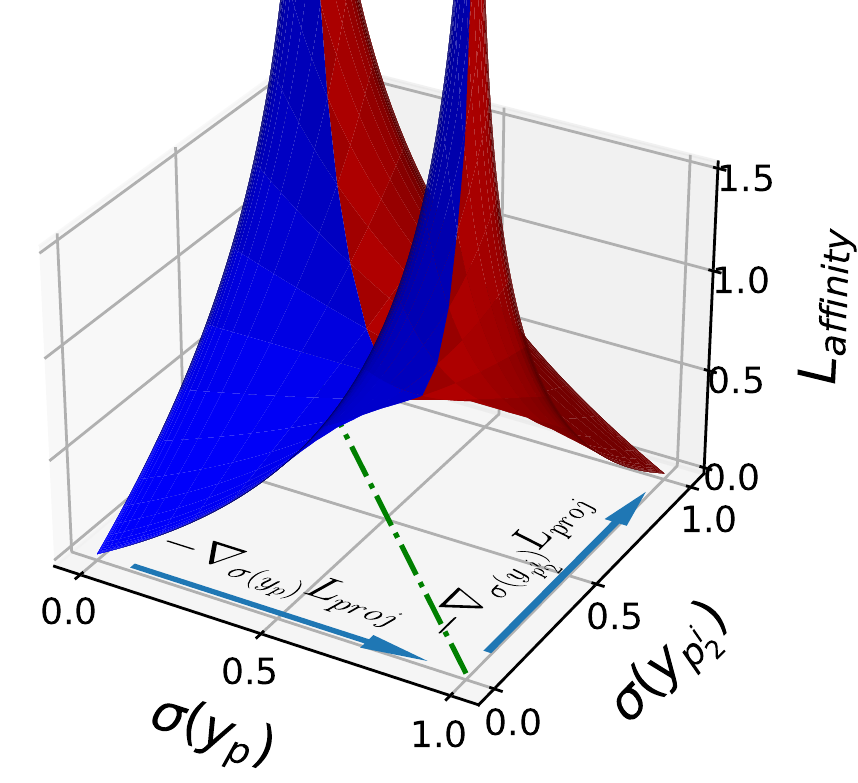}
        \caption{Symmetric Affinity Loss.}
        \label{fig:sym_two_halves_3d}
    \end{subfigure}
    \hfill
    \begin{subfigure}{0.49\linewidth}
        \includegraphics[width=\linewidth]{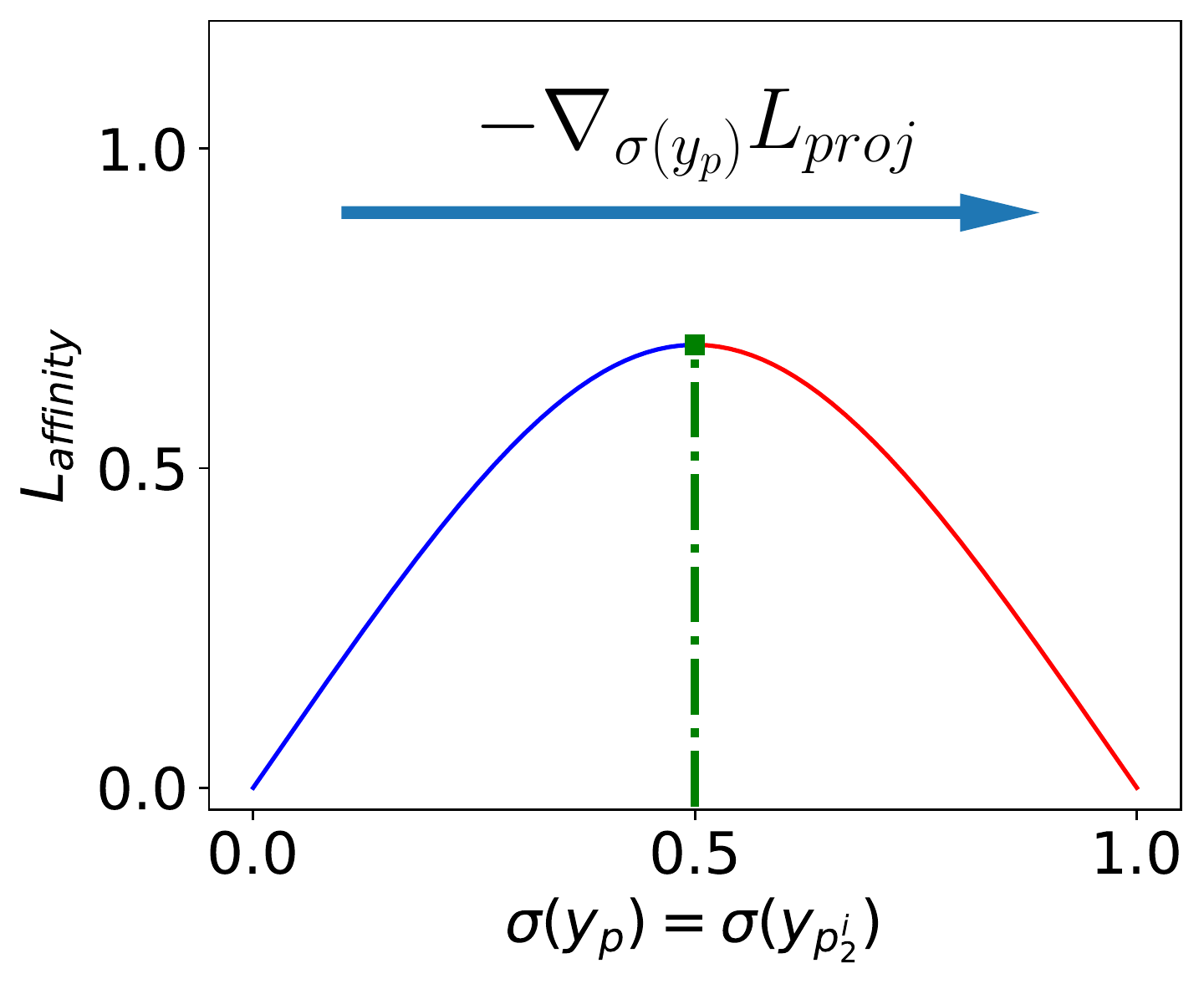}
        \caption{A 2d cross section of \cref{fig:sym_two_halves_3d}.}
        \label{fig:sym_two_halves_2d}
    \end{subfigure}
    \caption{\textbf{(a)} shows the saddle-shaped symmetric affinity loss in 3D in which horizontal axes are the probability of $p$ and $p_2^i$ being the foreground pixel. \textbf{(b)} shows the {\color{red}cross section} of symmetric affinity loss $\sigma(y_{p})=\sigma(y_{p_2^i})$. The landscape can be divided into two halves each of which converges to the double-negative location (\textcolor{Blue}{blue}) or the double-positive location (\textcolor{Red}{red}). Although the \textcolor{Green}{separation} between two halves is located in the middle, as the projection loss continuously encourages false positive pixels inside boxes, the \textcolor{Blue}{first half} of symmetric affinity loss is rendered less effective while the \textcolor{Red}{other half} accelerates convergence to trivial predictions.}
    \label{fig:sym_two_halves}
\end{figure}

BoxInst~\cite{tian2021boxinst} introduces color affinity loss to provide finer supervision in compensation for the coarse supervision from the projection loss which encourages the model to produce false positive pixels leading to trivial predictions such as bbox-shaped masks, but it does little to alleviate this problem if not worsen it. As shown in the \cref{fig:sym_two_halves}, symmetric affinity loss encourages pixel pairs linked by edges to be double-positive pairs as well as double-negative pairs. However, under the influence of non-positive gradient from the projection loss which constantly tends to increase the probability score $\sigma(y_p)$, it is far more likely for a pixel pair to land on the \textcolor{Red}{double-positive} half of the affinity loss than the \textcolor{Blue}{double-negative} one, accelerating the process to produce trivial solutions.

This results in a significant performance drop when naively adopting depth-grad affinity regularization in the form of \cref{equ:sym_depth}. Although color affinity loss proposed in Boxinst improve the performance by a stable and substantial margin, it shares this flaw leading to its current sub-optimal performance. More analysis about this will be discussed in \cref{subsec:exp_perf_asy,subsec:exp_asy_trivial}.

\section{Asymmetric Pairwise Affinity Loss}
\label{sec:asy_affinity}

Since this harmful coupling process which leads to trivial predictions is due to the affinity loss being symmetric, we intend to tackle this optimization problem by adjusting the landscape of affinity loss thus introducing asymmetry, which can provide a penalty against trivial predictions. We hence introduce our simple yet effective solution.

\subsection{Introducing Asymmetry with $\delta$}
\label{subsec:delta}
An intuitive solution to address this flaw of the affinity loss is by making it asymmetric therefore increasing the chance of pixel pairs falling on the double-negative half to compensate for the existing bias from the projection loss towards positive predictions. We hence make a simple modification to the original affinity loss by adding an offset $\delta$ to the \cref{equ:edge_prob}, \ie

\begin{equation}
    \begin{split}
        P(Y_e = 1; \delta) = &\sigma(y_p - \delta) \sigma(y_{p_2^i} - \delta) +\\
                             &\sigma(-y_p + \delta)\sigma(-y_{p_2^i} + \delta).
    \end{split}
    \label{equ:edge_prob_asy}
\end{equation}.

\begin{figure}[!t]
    \centering
    \begin{subfigure}{0.49\linewidth}
        \includegraphics[width=\linewidth]{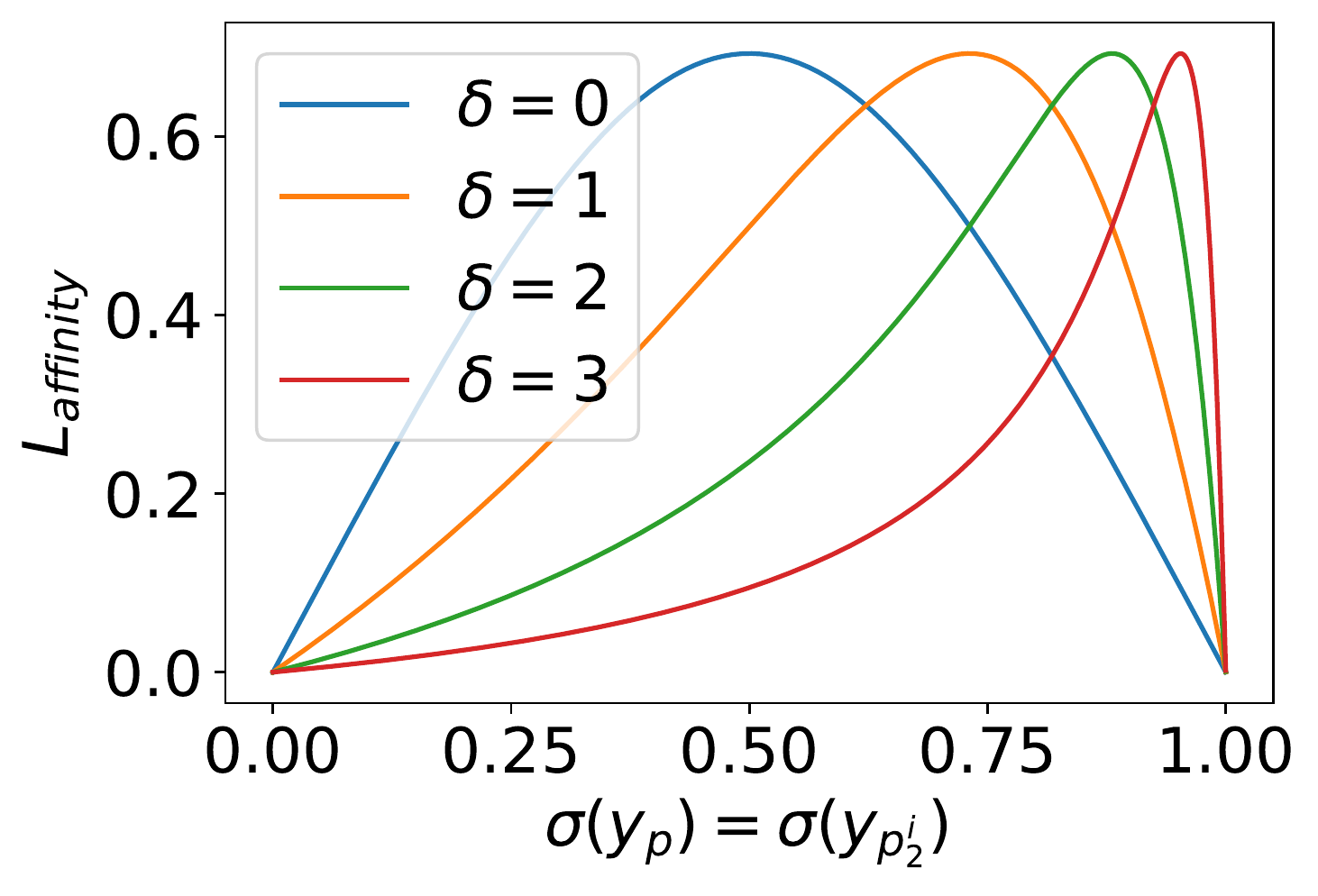}
        \caption{Symmetric Affinity ($\gamma = 0$).}
        \label{fig:short-delta}
    \end{subfigure}
    \hfill
    \begin{subfigure}{0.49\linewidth}
        \includegraphics[width=\linewidth]{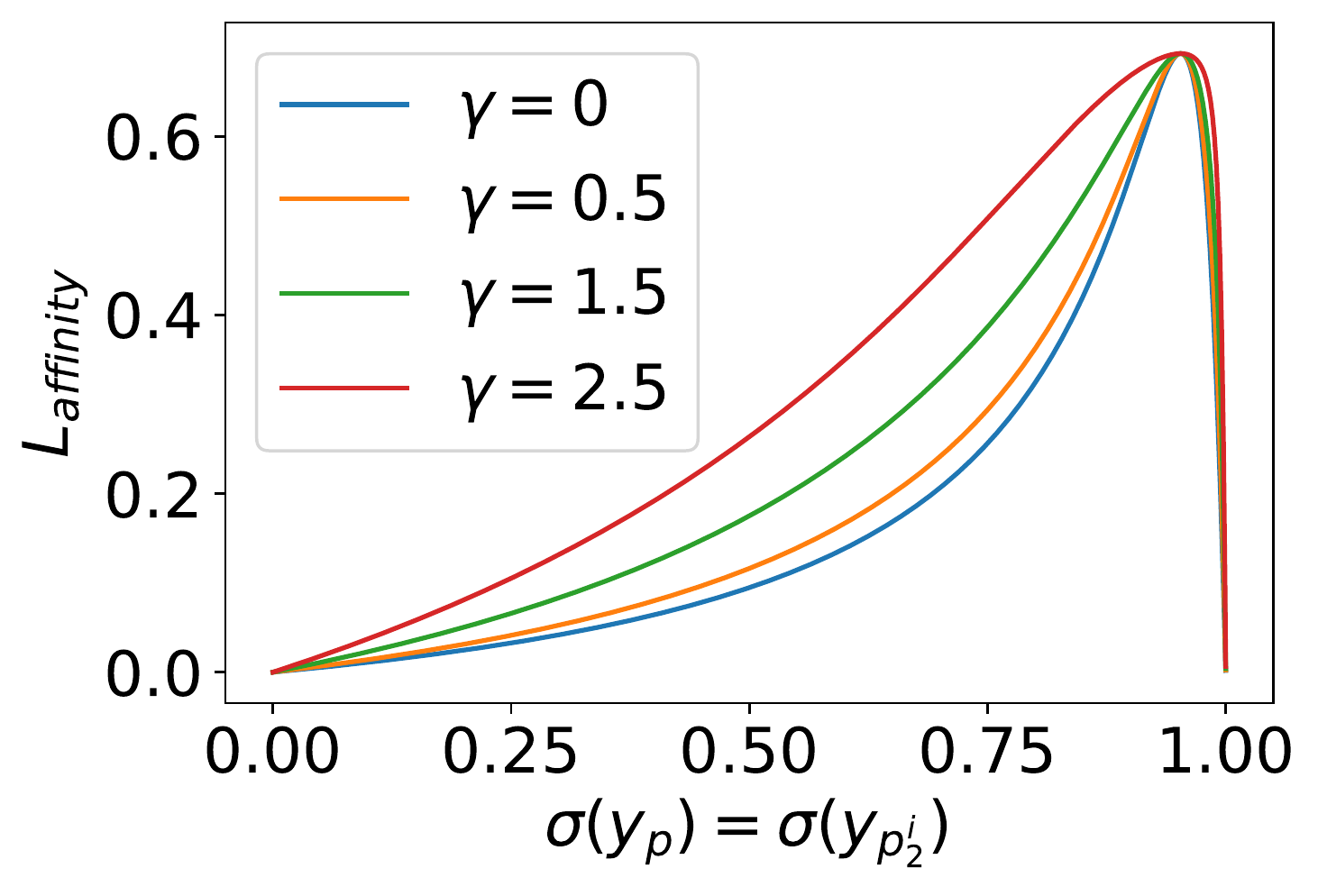}
        \caption{Asymmetric Affinity ($\delta = 3$).}
        \label{fig:short-gamma}
    \end{subfigure}
    \caption{The effect of $\delta$ on the shape of asymmetric affinity loss. It is shown in \textbf{(a)} that larger $\delta$ introduces stronger compensating bias but also slows convergence towards double-negative locations. As shown in \textbf{(b), introducing $\gamma$ can mitigate this issue.}}
    \label{fig:affinity_delta_gamma}
\end{figure}

As shown in \cref{fig:short-delta}, pairwise affinity loss with a positive $\delta$ is more likely to converge at $(\sigma(y_p), \sigma(y_{p_2^i})) = (0, 0)$ compared to $(1, 1)$ hence reduce the model's tendency to produce trivial predictions.

\subsection{Gradient Vanishing with Large $\delta$}
\label{subsec:gamma}
While introducing a strong bias, large $\delta$ results in a large plain area in the landscape of affinity loss causing gradient vanishing and slow convergence which undermine the purpose of using asymmetric affinity loss to introduce a bias towards negative predictions. We therefore propose to increase the gradient at a lower loss value by adding a modulating factor $e^{\gamma (P(Y_e = 1) - 0.5)}$ with a smoothness parameter $0 < \gamma < e \approx 2.71$ to the \cref{equ:sym_depth}. We then define the final formula of asymmetric depth-grad affinity loss as

\begin{equation}
% \begin{split}
    L_\text{depth} = -\frac{1}{N} \sum_{e \in E_\text{d}^\text{in}}^{\text{Diff}(e) \leq \tau_\text{d}}{e^{\gamma (P(Y_e = 1; \delta) - 0.5)}\log{P(Y_e = 1; \delta)}}.
% \end{split}
\label{equ:asy_depth}
\end{equation}

As illustrated in \cref{fig:short-gamma}, the proposed modulating factor in \cref{equ:asy_depth} greatly helps to alleviate the gradient vanishing issue caused by large $\delta$.

\begin{figure*}[!t]
    \centering
    \includegraphics[width=\linewidth]{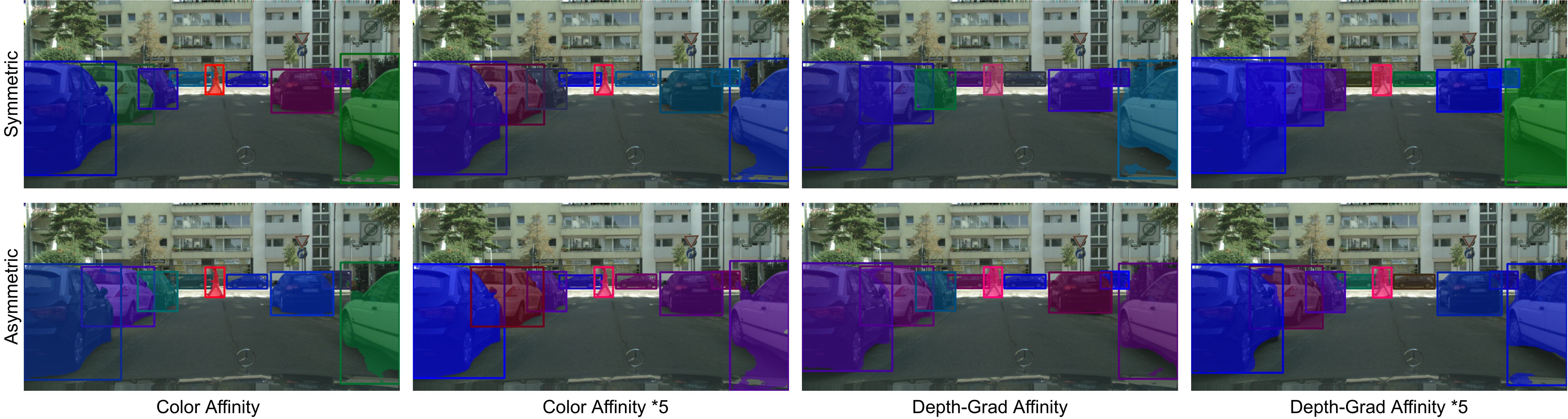}
    \caption{Qualitative results on Cityscapes validation set. Asymmetric affinity loss is shown to effectively prevent trivial predictions with both color and depth gradient affinity.}
    \label{fig:cityscapes}
\end{figure*}

\subsection{Overall Learning Objective}
\label{subsec:overall_objective}
The affinity loss functions for color and depth are defined in a similar fashion, therefore, we can propose a general form of asymmetric affinity loss as:
\begin{equation}
    \begin{split}
        &L_\text{affinity}(E^\text{in};S;\tau;\delta;\gamma) \\
        &= -\frac{1}{N} \sum_{e \in E^\text{in}}^{S(e) \geq \tau}{e^{\gamma (P(Y_e = 1; \delta) - 0.5)}\log{P(Y_e = 1; \delta)}}.
    \end{split}
    \label{equ:ays_affinity}
\end{equation}

Then the color affinity and depth affinity can be formulated as
\begin{equation}
    \begin{split}
        L_\text{color} &= L_\text{affinity}(E_c^\text{in};S_c;\tau_c;\delta_c;\gamma_c) \\
        L_\text{depth} &= L_\text{affinity}(E_d^\text{in};-\text{Diff};-\tau_d;\delta_d;\gamma_d)
    \end{split}
\label{equ:affinity_color_depth}
\end{equation}
where $E_c^\text{in}$ is the set of edges that at least has one pixel inside the box, and $S_c$ is color similarity threshold.

The overall learning objective for prediction in our \ours~is defined as
\begin{equation}
    L_\text{mask} = L_\text{proj} + \lambda_\text{c} L_\text{color} + \lambda_\text{d} L_\text{depth}
\label{equ:overall}
\end{equation}
where $\lambda_\text{c}$ and $\lambda_\text{d}$ as the respective loss weight for color and depth affinity loss which are linearly warmed up during the early iterations to stabilize the training. 

\begin{table}[!t]
    \centering
    \begin{tabular}{@{}lllllc@{}}
        \toprule
        \multirow{2}{*}{Method} & \multicolumn{2}{c}{col. affinity} & \multicolumn{2}{c}{dep. affinity} & \multirow{2}{*}{AP} \\
        \cmidrule(r){2-5}
        & Sym. & Asy. & Sym. & Asy. \\
        \midrule
        \textit{w/o color affinity:} \\
        Baseline & & & & & 18.596 \\
        \ours & & & $\checkmark$ & & 15.413\\
        \ours & & & & $\checkmark$ & \textbf{19.547} \\

        \midrule
        \textit{w/ color affinity:} \\
        Baseline & $\checkmark$ & & & & 21.145 \\
        \ours & $\checkmark$ & & $\checkmark$ & & 19.948\\
        \ours & $\checkmark$ & & & $\checkmark$ & 21.669 \\
        \ours & & $\checkmark$ & & $\checkmark$ & \textbf{24.687} \\
        \bottomrule
    \end{tabular}
    \caption{Ablation study of performance improvement from the asymmetric affinity loss compared to the symmetric one. Asymmetric Depth-Grad affinity loss performs significantly better than symmetric depth-grad affinity loss. Asymmetric color affinity loss also shows improvements compared to symmetric color affinity loss.}
    \label{tab:asymmetry}
    % \vspace{-2pt}
\end{table}

\begin{table*}[!t]
    \centering
    \begin{tabular}{@{}l|cc|cccccccc@{}}
        \toprule
        Method & AP & AP50 & person & rider & car & truck & bus & train & motor-cycle & bicycle \\
        \midrule
        BoxInst~\cite{tian2021boxinst} & 21.15 & 49.50 & 20.30 & 5.84 & 37.17 & 23.16 & 43.63 & 21.19 & 10.87 & 7.04 \\
        BoxLevelSet~\cite{li2022box} & 17.9 & 38.0 & 11.4 & 6.01 & 28.14 & 20.76 & 38.72 & 24.09 & 7.67 & 6.33 \\
        \ours~(w/o depth) & 24.10 & 52.27 & \textbf{23.29} & 9.33 & \textbf{39.32} & 26.57 & 43.47 & 27.14 & 13.60 & 9.37\\
        \ours & \textbf{24.69} & \textbf{53.01} & 22.91 & \textbf{9.89} & 39.24 & 26.38 & 46.62 & 31.89 & 12.52 & 8.05\\
        \bottomrule
    \end{tabular}
    \caption{Comparison with state-of-the-art methods on Cityscapes validation set. \ours{} achieves an improvement compared to previous methods even without depth supervision. By using depth supervision, \ours{} further improves the segmentation accuracy.}
    \label{tab:cityscapes}
\end{table*}

\section{Experiments}
\label{sec:experiments}

We conduct ablation studies about detailed designs in our model \ours~and compare \ours's performance with other state-of-the-art box-supervised instance segmentation methods on Cityscapes~\cite{Cordts2016Cityscapes}.

\subsection{Experimental Settings}

\paragraph{Dataset.} We evaluate our method on Cityscapes~\cite{Cordts2016Cityscapes}. Disparity maps from Cityscapes are converted to depth maps according to official instructions\footnote{\url{https://github.com/mcordts/cityscapesScripts/blob/master/README.md}}. Ablation studies are trained on Cityscapes training set and evaluated on Cityscapes validation set by default.

\paragraph{Model.} 
We use ResNet-50~\cite{he2016deep} with FPN~\cite{lin2017feature} as the backbone in all the experiments. The first stage of ResNet-50 and all the batchnorm layers are frozen following the default setting in Detectron2~\cite{wu2019detectron2}. We use BoxInst~\cite{tian2021boxinst} as the baseline model with or without symmetric color affinity loss. The depth gradient threshold $\tau_\text{d}$ is 0.01. $\delta_\text{c}$, $\gamma_\text{c}$, $\delta_\text{d}$, $\gamma_\text{d}$ are respectively 2.5, 1.5, 3.5, 2.5. Loss weights $\lambda_\text{c}$ and $\lambda_\text{d}$ are set as 1.0 and 0.1 respectively, which are linearly warmed up during the first 10k iterations. All the remaining hyper-parameters regarding affinity loss including color similarity threshold and dilation size are the same as ones in BoxInst unless specified.

\paragraph{Training Details.} Experiments on Cityscapes follow the training details in Detectron2\cite{wu2019detectron2} that models are trained with a total batch size of 8 on 8 GPUs (\ie NVIDIA RTX Titan) for 24k iterations using SGD optimizer with base learning rate set as 0.01 and reduced to 0.001 at iteration 18k. Weight decay and momentum for both datasets are set as 1e-4 and 0.9 respectively. Random flipping and scaling are applied sequentially for experiments on both Cityscapes. Due to the high resolution of images on Cityscapes, images are first randomly cropped to half size (512 * 1024) during training. No augmentation is applied during inference. Same as CondInst~\cite{tian2020conditional} which is the fully-supervised model our \ours~is based on, the stride of output segmentation masks is set to 2. BoxLevelSet~\cite{li2022box} in \cref{tab:cityscapes} is a re-implementation based on the offical code~\footnote{\url{https://github.com/LiWentomng/boxlevelset}} which does not support the Cityscapes dataset. To adapt to it, with some hyper parameter tuning, we set the learning rate as 1e-4.

\begin{table}[!t]
    \centering
    \begin{tabular}{@{}lcccc@{}}
        \toprule
        & \multirow{2}{*}{Sym.} & \multicolumn{3}{c}{$\delta_c(\delta_d)$} \\
        & & $1.5$ & $2.5$ & $3.5$ \\
        \midrule
        \textit{col. affinity:} & 21.145 & / & / & /\\
        $\gamma_c = 0.0$ & / & 22.868 & 23.144 & 17.927 \\
        $\gamma_c = 1.5$ & / & 23.821 & \textbf{24.095} & 23.410 \\
        $\gamma_c = 2.5$ & / & 24.076 & 24.062 & 23.933 \\
        \midrule
        \textit{dep. affinity:} & 15.413 & / & / & /\\
        $\gamma_c = 0.0$ & / & 18.127 & 18.308 & 18.653\\
        $\gamma_c = 1.5$ & / & 18.505 & 18.861 & 19.465\\
        $\gamma_c = 2.5$ & / & 18.379 & 19.069 & \textbf{19.547} \\
        \bottomrule
    \end{tabular}
    \caption{Performance of \ours~trained with different combinations of $\delta$ and $\gamma$. For better comparison, only color or depth gradient affinity loss is utilized during training. "Sym" represents symmetric affinity loss. A visualization of this table is shown in \cref{fig:delta_3d}.}
    \label{tab:affinity_delta_gamma}
\end{table}

\subsection{Performance Improvement from Asymmetry}
\label{subsec:exp_perf_asy}

Here we study how the introduced asymmetry in the affinity loss can reverse the negative impact of symmetric affinity loss with depth gradient affinity. We trained two sets of baseline models and \ours~with and without color affinity loss.

As shown in \cref{tab:asymmetry}, naively adopting symmetric affinity loss with depth gradient hurts performance regardless of whether color affinity is applied. However, the performance is improved in both cases after changing the depth gradient affinity loss from being symmetric to being asymmetric. We further discover that the performance of color affinity loss can still be improved by 3.018\% in mask AP when made asymmetric, proving that although the symmetric color affinity loss improves performance, it is sill sub-optimal due to its symmetry. These results indicate that our proposed asymmetric affinity loss can be used as a general form of affinity loss while the original symmetric affinity loss is only compatible with color affinity.

\subsection{Asymmetry Preventing Trivial Prediction}
\label{subsec:exp_asy_trivial}

As discussed in \cref{sec:proj_sym_affinity,sec:asy_affinity}, the original symmetric affinity loss accelerates convergence to trivial predictions while the asymmetric affinity in our \ours~prevents this. Here we study the effect of asymmetric affinity loss preventing trivial prediction with qualitative results on Cityscapes. To enhance the visual effect, we also conducted experiments with increased loss weight for color or depth gradient affinity loss to the 5 times of the original setting described in~\cref{subsec:overall_objective}, \ie $\lambda_\text{c}=5.0, \lambda_\text{d}=0.5$. To study the color and depth gradient affinity individually, color and depth gradient affinity loss are not utilized simultaneously during training in this ablation study.

The qualitative results shown in \cref{fig:cityscapes} demonstrate that asymmetry is the key to affinity loss to prevent trivial predictions, especially when utilized with depth gradient affinity. Mask predictions from models trained with symmetric affinity loss are usually squarer and fill more blank space inside boxes compared to models trained with asymmetric affinity.

\begin{figure}[!t]
    \centering
    \begin{subfigure}{0.49\linewidth}
        \includegraphics[width=\linewidth]{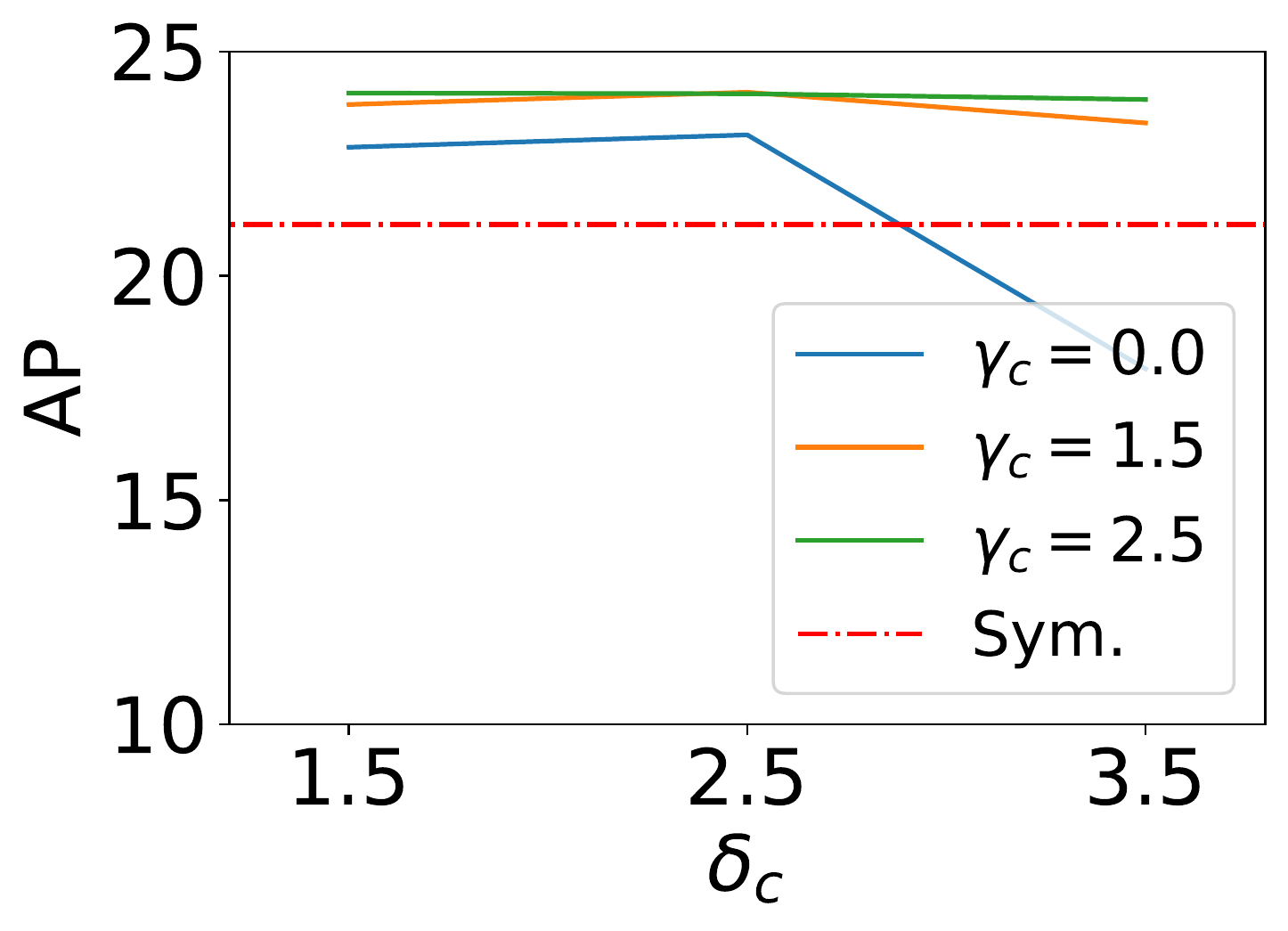}
        \caption{Color Affinity Only}
        \label{fig:short-a}
    \end{subfigure}
    \hfill
    \begin{subfigure}{0.49\linewidth}
        \includegraphics[width=\linewidth]{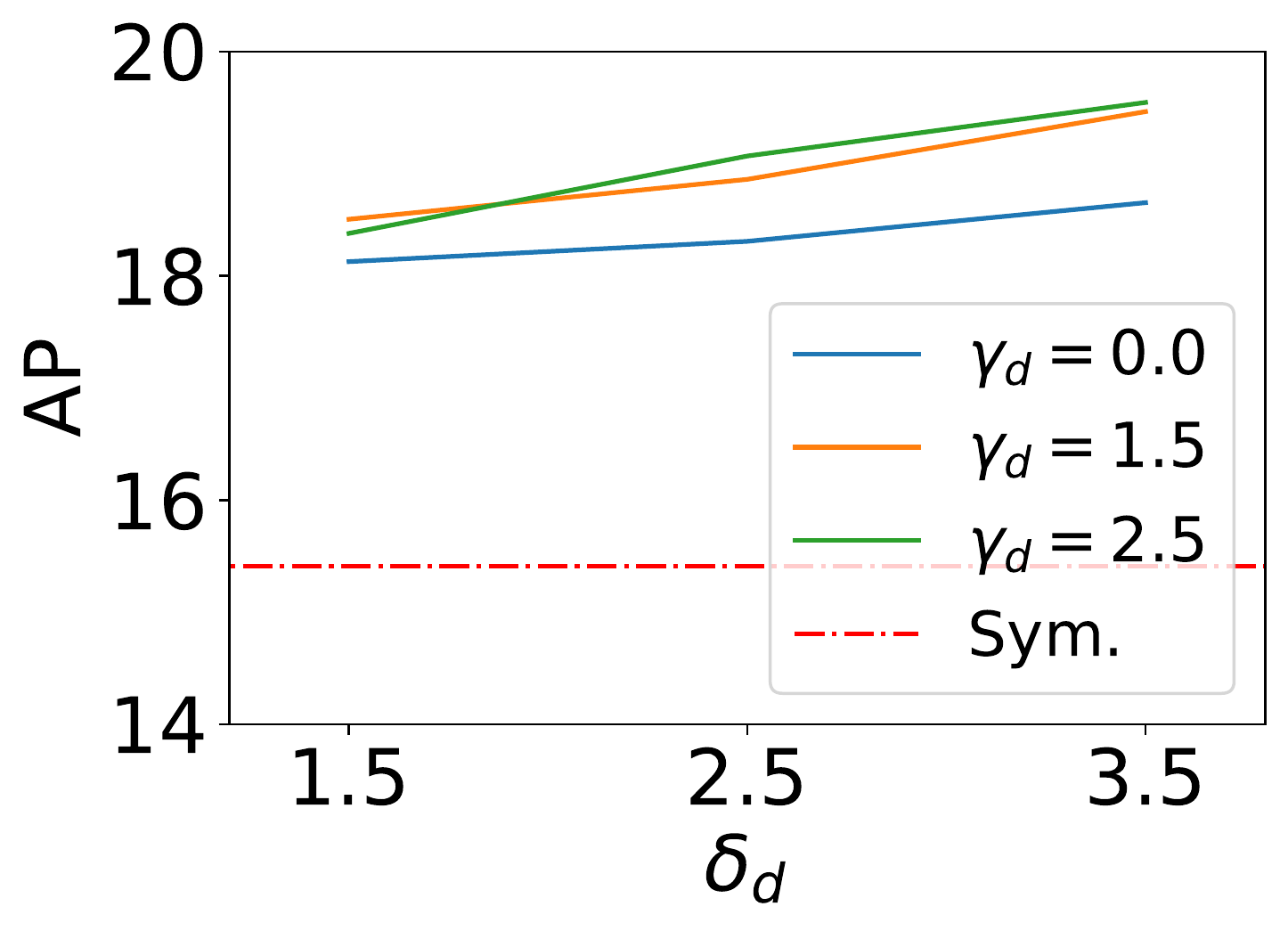}
        \caption{Depth Affinity Only}
        \label{fig:short-b}
    \end{subfigure}
    \caption{The ablation study about the effect of $\delta$ and $\gamma$ in asymmetric affinity loss with color and depth gradient affinity respectively. Large $\delta$ improves performance but may performance when being too large. Increasing $\gamma$ alleviates this performance drop.}
    \label{fig:delta_3d}
\end{figure}

\begin{figure*}[!t]
    \centering
    \includegraphics[width=\linewidth]{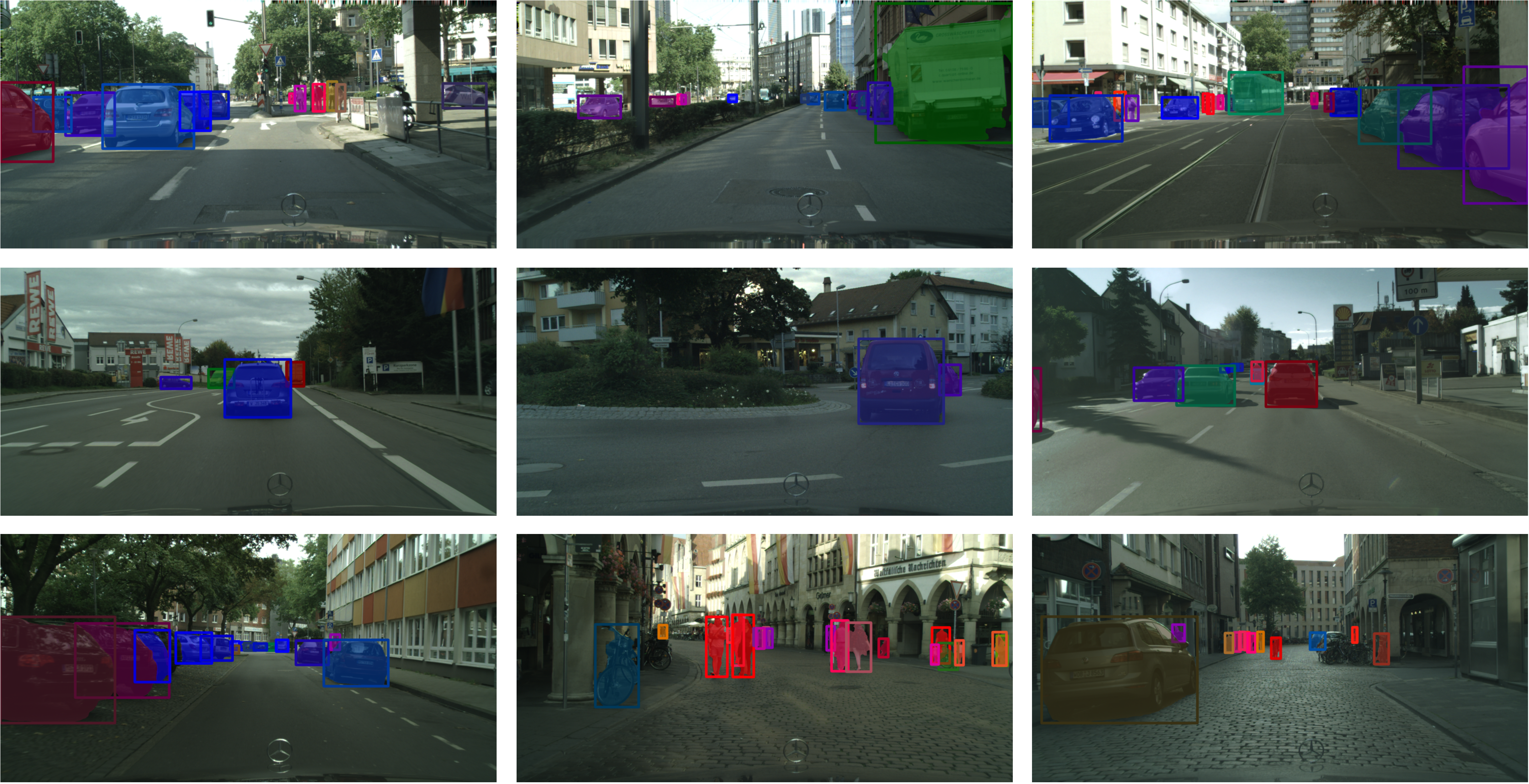}
    \caption{Some qualitative results showing the final performance of our \ours~on Cityscapes validation set.}
    \label{fig:final_cityscapes}
\end{figure*}

\subsection{Effect of $\delta$ and $\gamma$}
\label{subsec:exp_delta_gamma}

We introduce the $\delta$ in \cref{subsec:delta} and $\gamma$ in \cref{subsec:gamma} to control the affinity loss's asymmetry degree and gradient vanishing respectively. To examine whether these two hyper-parameters work as intended, we conduct the following ablation studies.

Performance of \ours~trained with different combinations of $\delta$ and $\gamma$ are shown in \cref{tab:affinity_delta_gamma}. For better comparison, we only use color affinity or depth gradient affinity for each experiment in this ablation study. As shown in \cref{fig:delta_3d}, performance tends to get improved with larger $\delta$ which enhances asymmetry and introduces stronger penalties against trivial predictions. However, extremely large $\delta$ could hurt performance due to gradient vanishing as discussed in \cref{subsec:gamma} while increasing $\gamma$ mitigates this performance drop.

\subsection{Comparison with State-of-the-art}

We compare \ours{} with state-of-the-art box-supervised instance segmentation methods on Cityscapes~\cite{Cordts2016Cityscapes}. As shown in \cref{tab:cityscapes}, \ours~outperforms all the previous methods by a noticeable margin with or without depth information, demonstrating the superiority of the proposed methods. \ours~without depth affinity still surpasses the BoxInst~\cite{tian2021boxinst} which is also our baseline model, by 3.09\% in mask AP. With additional depth information, \ours~outperforms BoxInst by a significant margin of 3.54\% in mask AP. It should also be noted that the BoxLevelSet is based on a stronger fully-supervised model that is SOLOv2~\cite{wang2020solov2} while \ours~is based on CondInst~\cite{tian2020conditional}.

Some qualitative results from our~\ours~with the highest performance on Cityscapes validation set are illustrated in \cref{fig:final_cityscapes}.

\section{Conclusion}
In this work, we reveal that the coupling interaction during the optimization between the projection loss and the affinity loss can lead to trivial box-shaped mask predictions when affinity loss is symmetric. This flaw of symmetry has a negative impact on the performance of affinity loss and makes it totally incompatible with other potentially helpful modalities such as depth gradients. Our proposed asymmetric affinity loss tackles this issue and can act as a general form of affinity loss with both color and depth gradient affinity. We believe this work can inspire future exploration into pairwise affinity regularization with various modalities.

%%%%%%%%% REFERENCES
{\small
\bibliographystyle{ieee_fullname}
\bibliography{egbib}

\begin{thebibliography}{10}\itemsep=-1pt

\bibitem{bolya2019yolact}
Daniel Bolya, Chong Zhou, Fanyi Xiao, and Yong~Jae Lee.
\newblock Yolact: Real-time instance segmentation.
\newblock In {\em Proceedings of the IEEE/CVF international conference on
  computer vision}, pages 9157--9166, 2019.

\bibitem{chen2020blendmask}
Hao Chen, Kunyang Sun, Zhi Tian, Chunhua Shen, Yongming Huang, and Youliang
  Yan.
\newblock Blendmask: Top-down meets bottom-up for instance segmentation.
\newblock In {\em Proceedings of the IEEE/CVF conference on computer vision and
  pattern recognition}, pages 8573--8581, 2020.

\bibitem{chen2019hybrid}
Kai Chen, Jiangmiao Pang, Jiaqi Wang, Yu Xiong, Xiaoxiao Li, Shuyang Sun,
  Wansen Feng, Ziwei Liu, Jianping Shi, Wanli Ouyang, et~al.
\newblock Hybrid task cascade for instance segmentation.
\newblock In {\em Proceedings of the IEEE/CVF Conference on Computer Vision and
  Pattern Recognition}, pages 4974--4983, 2019.

\bibitem{cheng2020boundary}
Tianheng Cheng, Xinggang Wang, Lichao Huang, and Wenyu Liu.
\newblock Boundary-preserving mask r-cnn.
\newblock In {\em European conference on computer vision}, pages 660--676.
  Springer, 2020.

\bibitem{Cordts2016Cityscapes}
Marius Cordts, Mohamed Omran, Sebastian Ramos, Timo Rehfeld, Markus Enzweiler,
  Rodrigo Benenson, Uwe Franke, Stefan Roth, and Bernt Schiele.
\newblock The cityscapes dataset for semantic urban scene understanding.
\newblock In {\em Proc. of the IEEE Conference on Computer Vision and Pattern
  Recognition (CVPR)}, 2016.

\bibitem{de2017semantic}
Bert De~Brabandere, Davy Neven, and Luc Van~Gool.
\newblock Semantic instance segmentation with a discriminative loss function.
\newblock {\em arXiv preprint arXiv:1708.02551}, 2017.

\bibitem{gao2019ssap}
Naiyu Gao, Yanhu Shan, Yupei Wang, Xin Zhao, Yinan Yu, Ming Yang, and Kaiqi
  Huang.
\newblock Ssap: Single-shot instance segmentation with affinity pyramid.
\newblock In {\em Proceedings of the IEEE/CVF International Conference on
  Computer Vision}, pages 642--651, 2019.

\bibitem{he2017mask}
Kaiming He, Georgia Gkioxari, Piotr Doll{\'a}r, and Ross Girshick.
\newblock Mask r-cnn.
\newblock In {\em Proceedings of the IEEE international conference on computer
  vision}, pages 2961--2969, 2017.

\bibitem{he2016deep}
Kaiming He, Xiangyu Zhang, Shaoqing Ren, and Jian Sun.
\newblock Deep residual learning for image recognition.
\newblock In {\em Proceedings of the IEEE conference on computer vision and
  pattern recognition}, pages 770--778, 2016.

\bibitem{NEURIPS2019_e6e71329}
Cheng-Chun Hsu, Kuang-Jui Hsu, Chung-Chi Tsai, Yen-Yu Lin, and Yung-Yu Chuang.
\newblock Weakly supervised instance segmentation using the bounding box
  tightness prior.
\newblock In H. Wallach, H. Larochelle, A. Beygelzimer, F. d\textquotesingle
  Alch\'{e}-Buc, E. Fox, and R. Garnett, editors, {\em Advances in Neural
  Information Processing Systems}, volume~32. Curran Associates, Inc., 2019.

\bibitem{huang2019mask}
Zhaojin Huang, Lichao Huang, Yongchao Gong, Chang Huang, and Xinggang Wang.
\newblock Mask scoring r-cnn.
\newblock In {\em Proceedings of the IEEE/CVF conference on computer vision and
  pattern recognition}, pages 6409--6418, 2019.

\bibitem{Khoreva_2017_CVPR}
Anna Khoreva, Rodrigo Benenson, Jan Hosang, Matthias Hein, and Bernt Schiele.
\newblock Simple does it: Weakly supervised instance and semantic segmentation.
\newblock In {\em Proceedings of the IEEE Conference on Computer Vision and
  Pattern Recognition (CVPR)}, July 2017.

\bibitem{lan2021discobox}
Shiyi Lan, Zhiding Yu, Christopher Choy, Subhashree Radhakrishnan, Guilin Liu,
  Yuke Zhu, Larry~S Davis, and Anima Anandkumar.
\newblock Discobox: Weakly supervised instance segmentation and semantic
  correspondence from box supervision.
\newblock In {\em Proceedings of the IEEE/CVF International Conference on
  Computer Vision}, pages 3406--3416, 2021.

\bibitem{li2022box}
Wentong Li, Wenyu Liu, Jianke Zhu, Miaomiao Cui, Xian-Sheng Hua, and Lei Zhang.
\newblock Box-supervised instance segmentation with level set evolution.
\newblock In {\em European Conference on Computer Vision}, pages 1--18.
  Springer, 2022.

\bibitem{li2017fully}
Yi Li, Haozhi Qi, Jifeng Dai, Xiangyang Ji, and Yichen Wei.
\newblock Fully convolutional instance-aware semantic segmentation.
\newblock In {\em Proceedings of the IEEE conference on computer vision and
  pattern recognition}, pages 2359--2367, 2017.

\bibitem{lin2017feature}
Tsung-Yi Lin, Piotr Doll{\'a}r, Ross Girshick, Kaiming He, Bharath Hariharan,
  and Serge Belongie.
\newblock Feature pyramid networks for object detection.
\newblock In {\em Proceedings of the IEEE conference on computer vision and
  pattern recognition}, pages 2117--2125, 2017.

\bibitem{liu2017sgn}
Shu Liu, Jiaya Jia, Sanja Fidler, and Raquel Urtasun.
\newblock Sgn: Sequential grouping networks for instance segmentation.
\newblock In {\em Proceedings of the IEEE International Conference on Computer
  Vision}, pages 3496--3504, 2017.

\bibitem{liu2018path}
Shu Liu, Lu Qi, Haifang Qin, Jianping Shi, and Jiaya Jia.
\newblock Path aggregation network for instance segmentation.
\newblock In {\em Proceedings of the IEEE conference on computer vision and
  pattern recognition}, pages 8759--8768, 2018.

\bibitem{newell2017associative}
Alejandro Newell, Zhiao Huang, and Jia Deng.
\newblock Associative embedding: End-to-end learning for joint detection and
  grouping.
\newblock {\em Advances in neural information processing systems}, 30, 2017.

\bibitem{tian2020conditional}
Zhi Tian, Chunhua Shen, and Hao Chen.
\newblock Conditional convolutions for instance segmentation.
\newblock In {\em European conference on computer vision}, pages 282--298.
  Springer, 2020.

\bibitem{tian2021boxinst}
Zhi Tian, Chunhua Shen, Xinlong Wang, and Hao Chen.
\newblock Boxinst: High-performance instance segmentation with box annotations.
\newblock In {\em Proceedings of the IEEE/CVF Conference on Computer Vision and
  Pattern Recognition}, pages 5443--5452, 2021.

\bibitem{wang2020solo}
Xinlong Wang, Tao Kong, Chunhua Shen, Yuning Jiang, and Lei Li.
\newblock Solo: Segmenting objects by locations.
\newblock In {\em European Conference on Computer Vision}, pages 649--665.
  Springer, 2020.

\bibitem{wang2020solov2}
Xinlong Wang, Rufeng Zhang, Tao Kong, Lei Li, and Chunhua Shen.
\newblock Solov2: Dynamic and fast instance segmentation.
\newblock {\em Advances in Neural information processing systems},
  33:17721--17732, 2020.

\bibitem{waqas2019isaid}
Syed Waqas~Zamir, Aditya Arora, Akshita Gupta, Salman Khan, Guolei Sun, Fahad
  Shahbaz~Khan, Fan Zhu, Ling Shao, Gui-Song Xia, and Xiang Bai.
\newblock isaid: A large-scale dataset for instance segmentation in aerial
  images.
\newblock In {\em Proceedings of the IEEE Conference on Computer Vision and
  Pattern Recognition Workshops}, pages 28--37, 2019.

\bibitem{wu2019detectron2}
Yuxin Wu, Alexander Kirillov, Francisco Massa, Wan-Yen Lo, and Ross Girshick.
\newblock Detectron2.
\newblock \url{https://github.com/facebookresearch/detectron2}, 2019.

\end{thebibliography}
}

\appendix

\twocolumn[
\centering
{\Large
\vspace{1.5em}
\textbf{AsyInst: Asymmetric Affinity with DepthGrad and Color for Box-Supervised Instance} \\
\vspace{0.5em}
\textbf{Appendix} \\
\vspace{1.0em}
}
]

\maketitle

\section{The upper limit for $\gamma$}

\begin{figure}[!t]
    \centering
    \includegraphics[width=\linewidth]{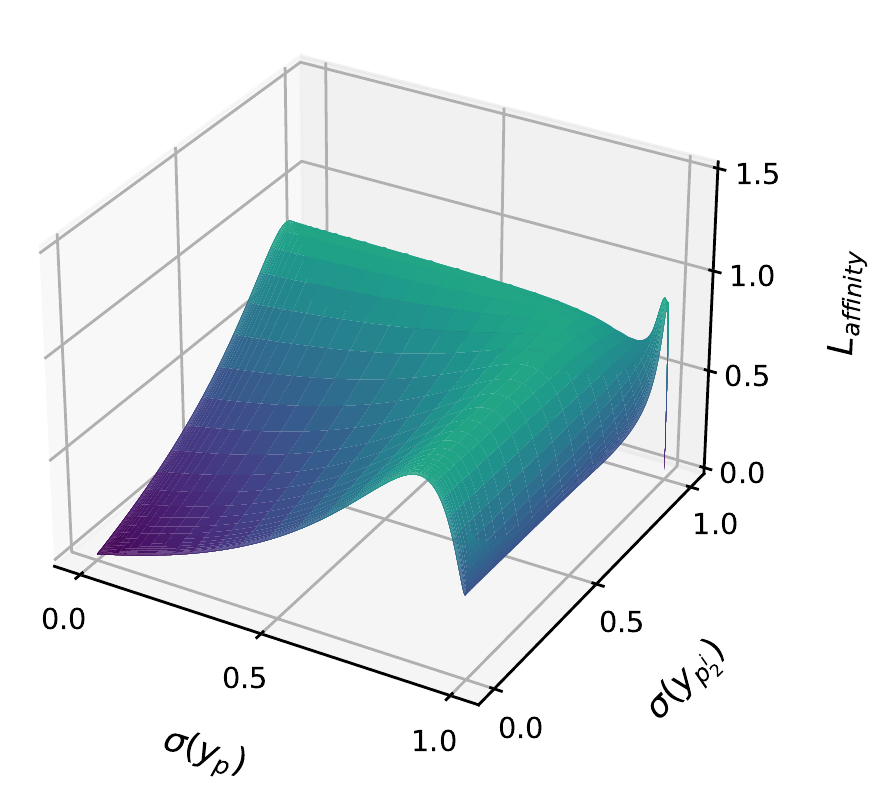}
    \caption{$\gamma$ being too large may create unwanted local minima. $\delta$ and $\gamma$ is set as 3.0 and 4.5 respectively in this figure.}
    \label{fig:supp_affinity_gamma_3d}
\end{figure}

\begin{figure}[!t]
    \centering
    \includegraphics[width=\linewidth]{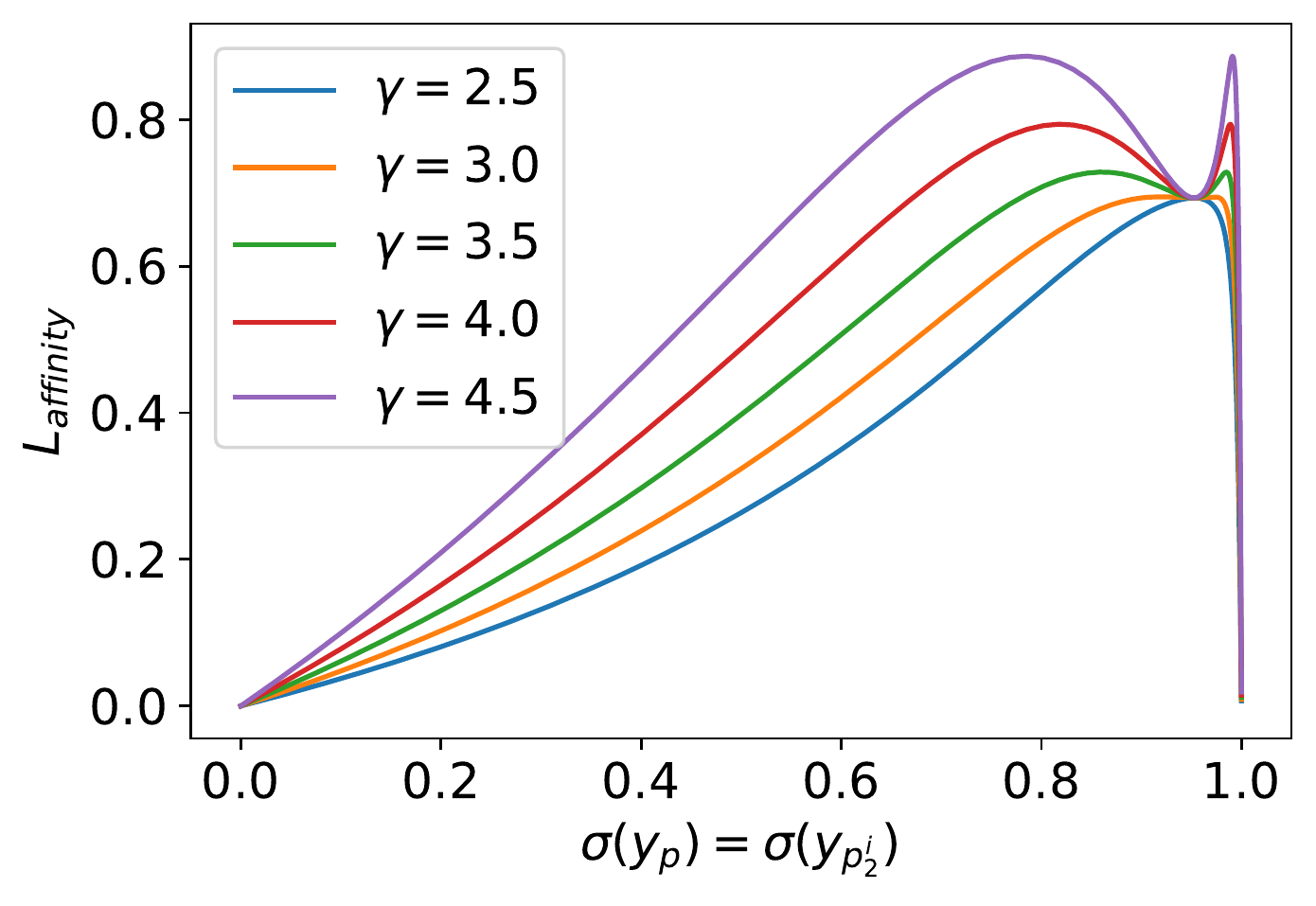}
    \caption{The effect of large $\gamma$ resulting in unwanted local minima. This figure only shows the cross section of affinity loss where $\sigma(y_p) = \sigma(y_{p_2^i})$ for clearer illustration.}
    \label{fig:supp_affinity_gamma}
\end{figure}

As illustrated in \cref{fig:supp_affinity_gamma_3d} and \cref{fig:supp_affinity_gamma}, a $\gamma$ ($\gamma > 0$) too large can increase the number of local maxima and make the previous maxima location another local minima, therefore there should be an upper limit for $\gamma$, \ie $0 < \gamma < e \approx 2.71$.
% \footnote{There is a typo in Sec. 5.2 that says $\gamma \leq \frac{2}{\ln{2}}$.}.

\subsection{Notations}
\label{subsec:notations}

Let $x_{p, \delta} = \sigma(y_p - \delta), 1 - x_{p, \delta} = \sigma(- y_p + \delta)$. Similarly, we have $x_{p_2^i, \delta} = \sigma(y_{p_2^i} - \delta), 1 - x_{p_2^i, \delta} = \sigma(- y_{p_2^i} + \delta)$. Specifically when $\delta = 0$, we define $x_{p} = \sigma(y_p), 1 - x_{p} = \sigma(- y_p), x_{p_2^i} = \sigma(y_{p_2^i}), 1 - x_{p_2^i} = \sigma(- y_{p_2^i})$. Also let $P_\delta = P(Y_e=1;\delta)$. Then we have

\begin{equation}
    \begin{split}
        L &= e^{\gamma (P_\delta - 0.5)}\log{P_\delta} \\
        P_\delta &= x_{p, \delta} x_{p_2^i, \delta} + (1 - x_{p, \delta}) (1 - x_{p_2^i, \delta}).
    \end{split}
\end{equation}

The gradient of sigmoid function $\sigma(\cdot)$ is
\begin{equation}
    \sigma'(y) = \frac{\partial \sigma(y)}{\partial y} = \frac{e^{-y}}{(1 + e^{-y})^{2}}.
\end{equation}

\begin{figure*}[!t]
    \centering
    \includegraphics[width=\linewidth]{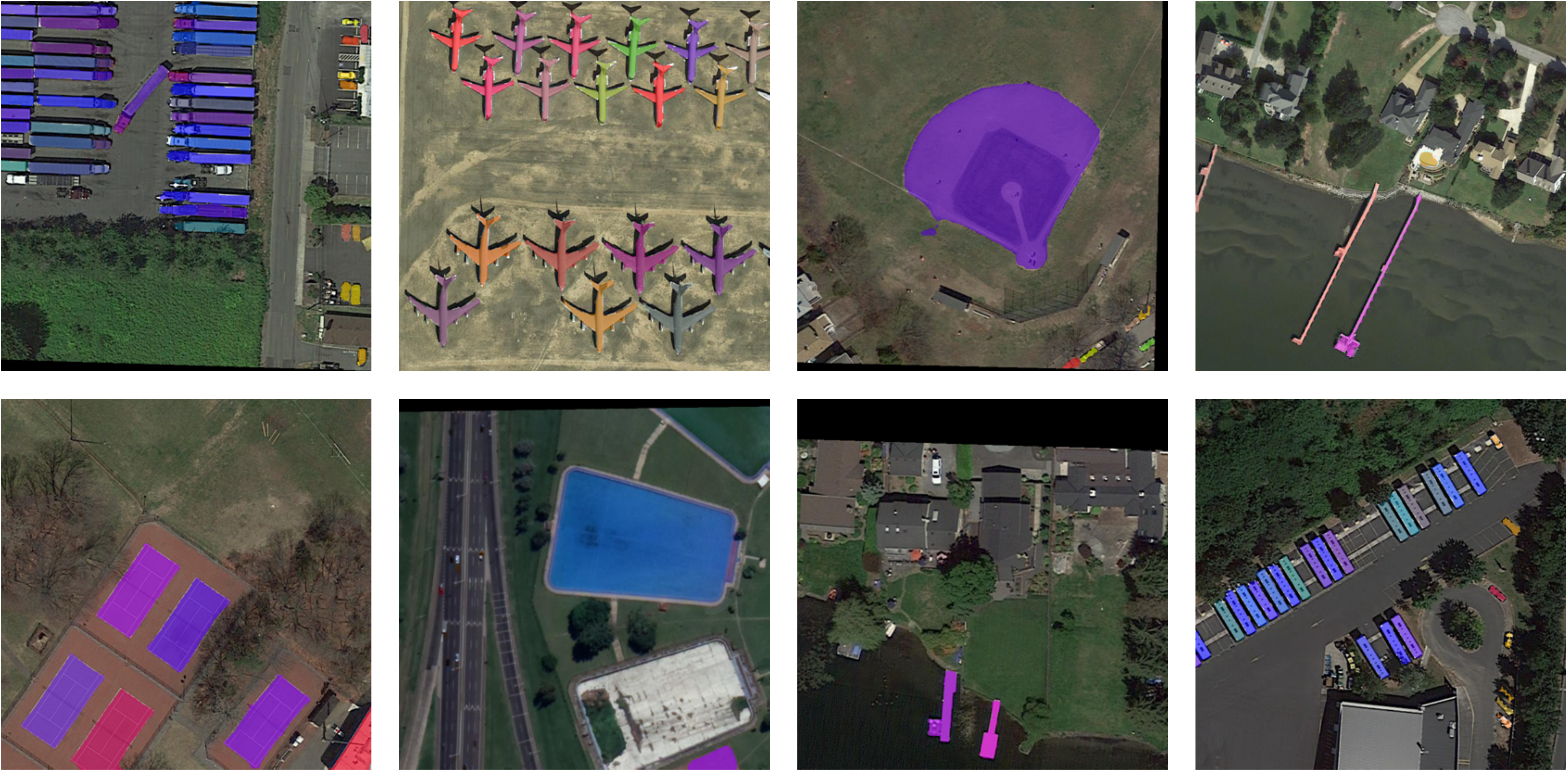}
    \caption{Some qualitative results from \ours~on iSAID.}
    \label{fig:supp_isaid}
\end{figure*}

\subsection{Proof}
\label{subsec:proof}

The affinity loss having only one maxima means that there is only one solution for         $\partial L/\partial \sigma(y_p) = 0$ and $\partial L / \partial \sigma(y_{p_2^i}) = 0$.

Considering $\partial L/\partial \sigma(y_p)$, we have

\begin{equation}
    \begin{split}
        \frac{\partial L}{\partial \sigma(y_p)} &= \frac{\partial L}{\partial x_p} \\
        &= \frac{\partial L}{\partial P_\delta} \frac{\partial P_\delta}{\partial x_{p, \delta}} \frac{\partial x_{p, \delta}}{\partial y_p} \frac{\partial y_p}{\partial x_{p}} \\
        &= \frac{\partial L}{\partial P_\delta} (2 x_{p_2^i, \delta} - 1) \sigma'(y_p - \delta) \frac{1}{\sigma'(y_p)}
    \end{split}
\end{equation}

where
\begin{equation}
    \begin{split}
        \frac{\partial L}{\partial P_\delta} &= e^{\gamma (P_\delta - 0.5)}\frac{\partial \log{P_\delta}}{\partial P_\delta} + \frac{\partial e^{\gamma (P_\delta - 0.5)}}{\partial P_\delta}\log{P_\delta} \\
        &= \frac{e^{\gamma (P_\delta - 0.5)}}{P_\delta} + \gamma e^{\gamma (P_\delta - 0.5)}\log{P_\delta} \\
        &= e^{\gamma (P_\delta - 0.5)}(\frac{1}{P_\delta} + \gamma\log{P_\delta})
    \end{split}
    \label{equ:dl_dp}
\end{equation}

Similarly, we have

\begin{equation}
    \begin{split}
        \frac{\partial L}{\partial \sigma(y_{p_2^i})} &= \frac{\partial L}{\partial x_{p_2^i}} \\
        &= \frac{\partial L}{\partial P_\delta} (2 x_{p, \delta} - 1) \sigma'(y_{p_2^i} - \delta) \frac{1}{\sigma'(y_{p_2^i})}
    \end{split}
\end{equation}

Obviously, one solution of

\begin{equation}
    \left\{
    \begin{split}
        \frac{\partial L}{\partial \sigma(y_p)} = 0\\
        \frac{\partial L}{\partial \sigma(y_{p_2^i})} = 0\\
    \end{split}
    \right.
    \label{equ:two_equation}
\end{equation}

is $(y_p, y_{p_2^i}) = (\delta, \delta)$, \ie $(x_{p, \delta}, x_{p_2^i, \delta}) = (0.5, 0.5)$.

Since the \cref{equ:two_equation} should have only one solution, $\partial L / \partial P_\delta = 0$, which is equivalent to $1/{P_\delta} + \gamma\log{P_\delta} = 0$ according to \cref{equ:dl_dp}, should have only one solution which is $(x_{p, \delta}, x_{p_2^i, \delta}) = (0.5, 0.5)$, or no solution at all. $P_\delta = 0.5$ when $(x_{p, \delta}, x_{p_2^i, \delta}) = (0.5, 0.5)$.

Let $f(P_\delta) = 1/{P_\delta} + \gamma\log{P_\delta}$. Then $f(P_\delta)$ must be always positive, or $\min{f(P_\delta)} = f(0.5) = 0$. $f(P_\delta)$ is at minima when $f'(P_\delta) = -1 / P_\delta^2 + \gamma / P_\delta$, \ie

\begin{equation}
    \begin{split}
        \argmin_{P_\delta}{f(P_\delta)} &= 1 / \gamma \\
        \min{f(P_\delta)} &= \gamma (1 - \log{\gamma})
    \end{split}
    \label{equ:argmin}
\end{equation}

Suppose $\min{f(P_\delta)} = f(0.5) = 0$, then $\gamma = 2 / \log{2}$, which contradicts with $\gamma = 2$ according to \cref{equ:argmin}.

Suppose $f(P_\delta)$ is always positive, then $\min{f(P_\delta)} = \gamma (1 - \log{\gamma}) > 0$, then we have $\gamma < e$.

To Summarize,

\begin{equation}
    0 < \gamma < e \approx 2.71
\end{equation}

\section{Experiments on iSAID}

The iSAID~\cite{waqas2019isaid} is a high-resolution remote sensing dataset with annotations for instance segmentation, featuring a large number of small objects with complex backgrounds.

We train \ours~and BoxInst~\cite{tian2021boxinst}, which is also our baseline, on iSAID for 12 epochs with only random horizontal flip as augmentation following BoxLevelSet~\cite{li2022box}. The learning rate is reduced by a scale factor of 0.1 at the 9-th and 11-th epoch. The number of proposals per image is restricted to 512 during training to reduce memory consumption. Other settings are the same as the ones used for experiments on Cityscapes~\cite{Cordts2016Cityscapes}.

\begin{table}[!t]
    \centering
    \begin{tabular}{@{}llc@{}}
        \toprule
        Method & backbone & mAP \\
        \midrule
        BoxInst & ResNet-50-FPN & 19.2 \\
        BoxLevelSet & ResNet-50-FPN & 20.1 \\
        \ours & ResNet-50-FPN & \textbf{20.2} \\
        \bottomrule
    \end{tabular}
    \caption{Comparison with SOTA on iSAID.}
    \label{tab:isaid}
\end{table}

\subsection{Comparison with State-of-the-art on iSAID}

As shown in \cref{tab:isaid}, \ours~achieves highest performance on iSAID compared to former SOTA BoxLevelSet~\cite{li2022box} while BoxLevelSet is built based on a stronger fully-supervised model that is SOLOv2~\cite{wang2020solov2}. Some qualitative results are shown in \cref{fig:supp_isaid}.

\subsection{Ablation study about $\delta$ and $\gamma$ on iSAID}

We also conducted a ablation study about the effect of $\delta$ and $\gamma$ on iSAID. Since iSAID doesn't provide depth maps, we only study $\delta$ and $\gamma$'s effect on color affinity here.

\begin{table}[!t]
    \centering
    \begin{tabular}{@{}lccccc@{}}
        \toprule
        % & \multirow{2}{*}{Sym.} & \multicolumn{3}{c}{$\delta_c(\delta_d)$} \\
        $\delta$ & Sym. & $0.5$ & $1.5$ & $2.5$ & $3.5$ \\
        \midrule
        % \textit{sym. col. affinity:}\\
        % 21.145\\
        % \textit{asy. col. affinity:}\\
        \textit{col. affinity:} & 19.2 & / & / & / & /\\
        % $\gamma = 0.0$ & / & 19.7 &  &  &  \\
        $\gamma = 1.5$ & / & 19.5 & 19.8 & 19.3 & 14.9 \\
        $\gamma = 2.5$ & / & 19.7 & \textbf{20.2} & 20.0 & 19.2 \\
        \bottomrule
    \end{tabular}
    \caption{Performance of \ours~trained with different combinations of $\delta$ and $\gamma$. "Sym." represents symmetric affinity loss.}
    \label{tab:isaid_ablation}
\end{table}

As shown in \cref{tab:isaid_ablation}, these results is very similar to ones in Tab. 3 as discussed in Sec. 6.4. It can be observed that asymmetry improves the performance of color affinity loss but extremely large $\delta$ may hurt performance. Increasing $\gamma$ may help mitigating this performance drop.

\end{document}